\documentclass[10pt,draftcls,peerreviewca,onecolumn]{IEEEtran} 

\usepackage{graphicx}
\usepackage{epstopdf}
\usepackage{float}
\usepackage{subfig}

\usepackage{cite}
\usepackage{amsmath, amssymb, mathbbol,color}
\usepackage[ruled,vlined]{algorithm2e}
 
\renewcommand{\QED}{\QEDopen}
\DeclareMathOperator*{\argmax}{arg\,max}

\usepackage[active]{srcltx} 
\usepackage{color}
\usepackage{rotating}
\usepackage[colorlinks=true,linkcolor=blue,citecolor=blue]{hyperref}
\usepackage{Milancommands}

\begin{document}
\title{Data-Driven Representations for Testing Independence: Modeling, Analysis and Connection with Mutual Information Estimation}
\author{Mauricio E. Gonzalez,  Jorge F. Silva, Miguel Videla, and Marcos E. Orchard
\thanks{M. Gonzalez, J. F. Silva, M. Videla, and M. Orchard are with the Information and Decision System Group, Department of Electrical Engineering,  University of Chile, Av. Tupper 2007 Santiago, 412-3, Room 508, Chile, Tel: 56-2-9784090, Fax: 56-2 -6953881,   (email: josilva@ing.uchile.cl).}
}

\maketitle

\begin{abstract}
This work addresses  testing the independence of two continuous and finite-dimensional random variables from the design of a data-driven partition. The empirical log-likelihood statistic is adopted to approximate the sufficient statistics of an oracle test against independence (that knows the two hypotheses).   It is shown that approximating the sufficient statistics of the oracle test  offers a learning criterion for designing a data-driven partition that connects with the problem of  mutual information estimation. Applying these ideas in the context of a data-dependent tree-structured partition (TSP), we derive conditions on the TSP's parameters to achieve a strongly consistent  distribution-free test of independence over the family of probabilities equipped with a density.  Complementing this result, we present finite-length results that show our TSP scheme's capacity to detect the scenario of independence structurally with the data-driven partition as well as new sampling complexity bounds for this detection.  Finally, some experimental analyses provide evidence regarding our scheme's advantage for testing independence  compared with some strategies that do not use data-driven representations.
\end{abstract}


 \newtheorem{corollary}{\bf COROLLARY}
\newtheorem{theorem}{\bf THEOREM}
\newtheorem{lemma}{\bf LEMMA}
\newtheorem{proposition}{\bf PROPOSITION}
\newtheorem{definition}{\bf Definition}
\newtheorem{remark}{\bf Remark}

\section{Introduction}
\label{sec_intro}
The problem of detecting independence from i.i.d. samples in a learning setting (distribution-free) 
is fundamental in statistics and has found numerous applications in statistical signal processing, 
data analysis, machine learning, inference, and decision problems \cite{ku_2005,common_1994,cardoso_2001,chen_2009}. 
For instance, the capacity to detect independent and conditional independent structures from data is relevant when having invariances and probabilistic symmetries in decision and machine learning models \cite{Bloem_2019, runge_2018, bellot_2019,Sen_2017}. This capacity has been used in blind source separation,  independent component analysis (ICA) \cite{ku_2005,common_1994,cardoso_2001,chen_2009,ku_2006} 
and the detection of associations in data \cite{reshef_2011}.
%
Detecting independence from data has been systematically studied, and the literature is vast. Different non-parametric strategies have been proposed for this task using kernel-based statistics \cite{gretton_2005,gretton_2008,gretton_2007,pfister_2016}, distance-based approaches \cite{szekely_2007,szekely_2009,szekely_2013,zheng_2012,wang_2017}, histogram-based approaches \cite{gretton_2010, ma_2019, zhang_2019,reshef_2011,heller_2016a,heller_2016b}, 
log-likelihood statistics  \cite{gretton_2010,menedez_2006}, correlation-based approaches \cite{han_2017,dauxois_1998}, $\phi$-divergence estimators \cite{menedez_2006,gretton_2010}, entropy and mutual information estimators \cite{dionisio_2006,Suzuki_2016, kontoyiannis_2016, reshef_2011,reshef_2011}, among many others.

A natural strategy, and the one we follow in this paper, is to partition the observation space (the binning approach) to evaluate the discrepancy (in some sense) between two empirical distributions \cite{gretton_2010,gretton_2008,Suzuki_2016, szekely_2007,zhang_2019,ma_2019}. We highlight the following work using this approach: 
Gretton and Gy\"{o}rfi \cite{gretton_2010} introduced a family of product data-independent partitions using the $L_1$-statistics \cite{devroye_2001} and the I-divergence statistics \cite{kullback1958}. 
With these two approaches, 
sufficient conditions are established on the partition and other learning parameters to achieve strong consistency for detecting independence distribution-free in a continuous multivariate scenario. 
Szekely {\em et al.} \cite{szekely_2007} also use partitions on the sample space, utilizing empirical distance correlation and distance covariance statistics to introduce a novel test of independence. The test is consistent and
shows better empirical results than classical likelihood-ratio tests in scenarios with non-linear dependencies. 
Shi {\em et al.} \cite{shi2020distribution} address the multivariate scenario by incorporating the concept of  transportation-based ranks and signs. They propose plugging the center-outward distribution function into the mentioned distance covariance statistic \cite{szekely_2007} to determine a distribution-free test of independence.

More recently, Zhang \cite{zhang_2019} proposed a non-parametric test of independence assuming a rich class of models with known marginal distributions (uniform over $[0,1]$). This work proposed a binary expansion filtration of the space and deeply studied some symmetric properties to create a novel test that is uniformly consistent (on the power loss) and minimax for the power (in the sample size) assuming a large class of alternative distributions. 
Zhang {\em et al.} \cite{zhang2021beauty} presented a novel extension of the binary expansion filtration strategy \cite{zhang_2019} to a multidimensional setting by cleverly approximating an oracle {\em Neyman-Pearson} (NP) statistic. 
They propose a  data-adaptive weight strategy (for the binary expansion) to approximate the ideal weight of the NP test. This new test unifies several important tests ($\chi^2$ test statistics, distance correlation) and shows promising empirical results.

On the important connection between testing independence and
information measures \cite{csiszar_2004,cover_2006}, we highlight the work of Kontoyiannis and Skoularidou \cite{kontoyiannis_2016}, where the problem of estimating directed information (directed information rate) between two discrete (finite alphabet) processes was explored based on a plug-in estimator.  Importantly, the authors used this plug-in estimator to test the presence or absence of causality between the processes. Along the same lines, Suzuki \cite{Suzuki_2016} explored different strategies to estimate mutual information between discrete and continuous random variables and then evaluated those strategies experimentally for the task of testing independence. 
 Finally, Reshef {\em et al.} \cite{reshef_2011} introduced the maximal information coefficient (MIC) to measure complex functional dependencies between a pair of variables by maximizing the (empirical) mutual information computed for collections of (adaptive) partitions of different resolutions.
 
 In many of the methods mentioned that use partitions (or binning) for testing independence, a component that has not been explored systematically is the role played by data-driven partitions \cite{lugosi_1996_ann_sta,darbellay_1999,reshef_2011}. 
Data-driven partitions use data for the binning process (vector quantization) to construct the cells and define the final structure of the partition.  This flexibility offers the capacity to better address inference and learning problems.  Supporting this idea, data-driven representations have shown great approximation properties and better decision-estimation performance in numerous non-parametric learning problems \cite{lugosi_1996_ann_sta, darbellay_1999, silva_2012, silva_2010, silva_2010b}.

In our context, the motivation that piqued our interest in data-driven partitions is the fact that under the hypothesis of independence, the trivial partition (the partition with one cell that contains all the space) is a sufficient representation of the problem. Our initial conjecture is that a well-designed adaptive data-driven partition could have the ability to detect (learn from data) this trivial solution by implementing a form of explicit regularization in its design. This adaptive ability could provide better detection of independence than non-adaptive partition schemes, which has prompted the exploration of data-driven methods for testing independence in this work.

Motivated by this, we look at the problem of designing a new learning criterion 
that selects a data-driven partition of the space (adaptive binning)  as the optimal trade-off between estimation and approximation errors in a learning problem.
To formulate this representation learning task,  we adopt ideas from universal source coding to introduce a regret term that measures the discrepancy attributed to the use of empirical log-likelihood statistics --- restricted over a partition --- with respect to the oracle sufficient statistics of an ideal  test that knows the true (two) probabilities: the oracle test against independence \cite{cover_2006,blahut_1974}. Using this regret analysis,  we establish a novel connection with the problem of mutual information (MI) estimation \cite{kontoyiannis_2016}. Furthermore, general conditions are derived to obtain a strongly consistent test of independence in the strong (almost-sure and distribution-free) sense introduced in \cite{gretton_2010}. 

We apply this framework in the context of  tree-structured partitions (TSP)  \cite{silva_2012},  
which is the main application focus of this work.
 TSP algorithms were selected for the following important reasons: Their implementation is simple, TSP have a binary structure that can be used to address learning tasks efficiently for a large class of optimization problems (minimum cost-tree pruning) \cite{breiman_1984,scott_2006,scott_2005}, and they have been shown to be expressive (with good approximation properties) when compared with other partition strategies in non-parametric learning and decision tasks \cite{lugosi_1996_ann_sta, darbellay_1999, silva_2012, silva_2010, silva_2010b,breiman_1984,devroye1996,lugosi_1996_ann_sta,scott_2006}.

On the proposed test, we derive concrete connections with results on mutual information estimation \cite{silva_2012}. From these connections, we established new design conditions to obtain a strongly consistent test of independence and, 
more importantly, non-asymptotic results that express our framework's capacity to approximate the information term of the oracle (ideal) test with a finite sample-size.  Going a step further in this finite length performance analysis, we study our scheme's capacity to detect independence structurally with the underlying data-driven partition, which was  one of the original motivations used to explore data-driven partition for this task.

Indeed, we show that under the independence assumption, our data-driven partition collapses to the trivial solution with one cell (implying that the resulting MI estimator is zero) with a finite sample-size almost surely. This is a remarkable property attributed to our scheme's capacity to adapt its representation to the sufficient statistics of the problem, which is the trivial partition in the context of independence. From this ability, we improve our original result concerning our test's consistency (density-free) and provide refined sampling complexity bounds for detecting the scenario of independence. 

To implement our approach, we propose a learning algorithm that offers a computationally efficient implementation of our data-driven strategy for testing independence.  
The algorithm is divided in two important phases (a growing and pruning stage), which are both shown to have a polynomial complexity on the side of the problem (the sample size).
Finally, we provide empirical evidence of the advantage of our strategy in some controlled simulation scenarios. For this last analysis, we introduce concrete non-asymptotic metrics (in the form of sampling complexity indicators) to measure the test's ability to detect the correct hypothesis with a finite number of samples.   

A preliminary version of this work was presented in \cite{gonzales_2020}. This work significantly expands the technical contribution in \cite{gonzales_2020} and explores implementations and experimental analyses.

\subsection{Organization of the Paper}
Section \ref{sec_problem_st} introduces the problem and basic notations.
Section \ref{sec_plug_in_detector} presents the general histogram-based scheme, and Section \ref{sec_regret} introduces the regret-based analysis proposed in this work and its connection with the problem of MI estimation. Section \ref{sub_sec_regret_consistency} presents a general consistency result  for our problem  (Theorem \ref{th_main}). Section \ref{sec_tsp} 
introduces the family of tree-structured data-driven partitions and  presents consistency results (Theorem \ref{th_nearly_minimax_opt_regret}, Lemma \ref{th_consistency_regret} and  Corollary \ref{th_test_consistency}). Section \ref{sec_finite_length} analyzes some finite-length properties (Theorem \ref{th_nearly_minimax_opt_regret}) and
elaborates on the structural capacity of our scheme to detect independence under the null hypothesis (Lemma \ref{th_structural_detection_ind}, and  Theorems  \ref{th_test_consistency_2} and \ref{th_bound_on_tail_of_T0}). 
Sections \ref{sec_implementation} and \ref{sec_implementation_simulations} cover implementation and empirical analysis, respectively. Finally, Section \ref{sec_final} concludes with some discussion and future directions.  The proofs are presented in the Appendix.

\section{Preliminaries}
\label{sec_problem_st}
Let us consider two random variables $X$ and $Y$ taking values in $\mathbb{R}^p$ and $\mathbb{R}^q$, respectively,  with joint distribution $P$ in $(\mathbb{R}^d, \mathcal{B}(\mathbb{R}^d))$, where $d=p+q$ and  $\mathcal{B}(\mathbb{R}^d)$ is the Borel sigma field. Let us denote by $\mathcal{P}_0$ the class of probabilities in $(\mathbb{R}^d, \mathcal{B}(\mathbb{R}^d))$ under which $(X,Y)$ are independent, meaning that if $P\in \mathcal{P}_0$, then $P(A \times B)= P(\mathbb{R}^p \times B) \cdot P(A \times \mathbb{R}^q)$ for any $A\in  \mathcal{B}(\mathbb{R}^p)$ and $B\in  \mathcal{B}(\mathbb{R}^q)$.

Given $n$ i.i.d. samples of $(X,Y)$,  denoted by $Z^n_1 \equiv (Z_1\equiv(X_1,Y_1), Z_2\equiv (X_2,Y_2),..,Z_n \equiv(X_n,Y_n))\in \mathbb{R}^d$ driven by the model $P\in \mathcal{P}(\mathbb{R}^d)$,  the problem is to decide from $Z^n_1$ whether $X$ and $Y$ are independent (meaning that $P\in \mathcal{P}_0$) or, alternatively, $X$ and $Y$ have some statistical dependency, i.e., $P\notin \mathcal{P}_0$. 
In this context, a decision rule $\phi_n(\cdot)$ of length $n$ is a function from $(\mathbb{R}^{d})^n$ to $\left\{0,1\right\}$,  where  $\phi_n(z^n_1)=0$ means that the rule decides that the underlying probability producing $z^n_1$ belongs to $\mathcal{P}_0$. 
Then, for any decision rule $\phi_n\in \Pi_n$\footnote{$\Pi_n$ denotes the collection of binary rules acting on $\mathbb{R}^{dn}$.} of length $n$, we recognize two errors:
\begin{itemize}
\item Assuming that $P\in \mathcal{P}_0$ ($\mathcal{H}_0$),  the  non-detection of independence between $X$ and $Y$ is measured by\footnote{$\mathbb{P}$ denotes the entire process distribution of $(Z_n)_{n\geq 1}$ and $P^n$ is the $n$-fold probability in $\mathbb{R}^{d n}$ induced by the marginal $P$.} 
$$\mathbb{P}(\phi_n(Z^n_1)=1)=P^n(\left\{z^n_1: \phi_n(z^n_1)=1 \right\}),$$
\item Assuming that $P\notin \mathcal{P}_0$ ($\mathcal{H}_1$), the false detection of independence between $X$ and $Y$ is measured by 
$$\mathbb{P}(\phi_n(Z^n_1)=0)=P^n(\left\{z^n_1: \phi_n(z^n_1)=0 \right\}).$$
\end{itemize}

The following is the classical notion of strong-consistency used to evaluate the asymptotic goodness of a universal scheme (collection of 
rules of different lengths) for detecting independence from data \cite{gretton_2010}.
\begin{definition}\label{def_consistency}
Given a 
scheme $\xi=\left\{\phi_n \in \Pi_n, n\geq 1 \right\}$, where  $\phi_n$ is a decision rule of length $n$,  
we say that $\xi$ is strongly consistent for detecting independence if
\begin{itemize}
\item[i)] under $\mathcal{H}_0$: 
$(\phi_n(Z^n_1))_{n\geq 1}$ reaches $0$ eventually with probability one, or $\mathbb{P}$-almost surely (a.s.), 
\item[ii)]  under $\mathcal{H}_1$: $(\phi_n(Z^n_1))_{n\geq 1}$  reaches $1$ eventually $\mathbb{P}$-a.s.
\end{itemize}
\end{definition}

\subsection{The Divergence and Mutual Information}
Let $P$, $Q$ be two probability measures in $(\mathbb{R}^d, \mathcal{B}(\mathbb{R}^d))$
such that $P \ll Q$, and let us consider $\pi= \left\{A_i, i\in \mathcal{I} \right\}$  a measurable partition of $\mathbb{R}^d$ 
where $I$ is finite or countable.  The divergence of $P$ with respect to $Q$ restricted over
$\pi$ (or the sub-sigma field induced by $\pi$ denoted by $\sigma(\pi)$) is given by \cite{kullback1958,cover_2006}
\begin{equation} \label{eq_pre_1}
	D_{\sigma(\pi)}(P||Q) \equiv \sum_{i\in \mathcal{I}} P(A_i) \log \frac{ P(A_i)}{ Q(A_i)}. 
\end{equation}
The 
divergence of $P$ with respect to $Q$ is \cite{kullback1958,cover_2006}
\begin{equation}\label{eq_pre_2}
	D(P||Q) \equiv \sup_{\pi \in \mathcal{Q}(\mathbb{R}^d)}D_{\sigma(\pi)}(P||Q)
\end{equation}
where $\mathcal{Q}(\mathbb{R}^d)$ is the collection of measurable\footnote{$\mathcal{B}(\mathbb{R}^d)$ is the sigma field of Borel sets.} partitions of $\mathbb{R}^d$.

\section{The Empirical Log-Likelihood Statistics 
from a Data-Driven Partition}
\label{sec_plug_in_detector}
In this work, we adopt a histogram-based log-likelihood statistics approach \cite{gretton_2010}.  
We interpret this approach as an empirical version of the Neyman-Pearson (NP) test against independence \cite{blahut_1974}. 
The  ideal (oracle) NP test  against independence decides whether the samples belong to the following two known cases: $P$ (for some $P\notin \mathcal{P}_0$) or $Q^*(P)\in \mathcal{P}_0$,  where $Q^*(A \times B)= P(\mathbb{R}^p \times B) \cdot P(A \times \mathbb{R}^q)$ for any $A\in  \mathcal{B}(\mathbb{R}^p)$ and $B\in  \mathcal{B}(\mathbb{R}^q)$.\footnote{$Q^*(P)$ can be interpreted as the projection of $P$ on $\mathcal{P}_0$ in the information divergence sense, i.e., $Q^*(P)$ is the solution of $\min_{Q\in \mathcal{P}_0} D(P||Q)$.}  

The specific approach proposed in this work is a two-stage process that uses the data $Z^n_1\sim P^n$ twice: first, to estimate $P$,  and its projected version $Q^*(P)$ (over  $\mathcal{P}_0$), with the caveat that the estimation is restricted over the events of a finite partition of the joint space $\mathbb{R}^d$ (i.e., a histogram-based estimation of $P$ and $Q^*(P)$, respectively); second, to compute an empirical version of the log likelihood-ratio to decide  (using a threshold) if $\hat{P}=\hat{Q}^*$ (independence)  or $\hat{P} \neq \hat{Q}^*$ (non-independence). 

There are two elements of learning design that determine our empirical test.  The first is a data-driven partition rule denoted 
by $\pi_n(\cdot)$, which is a function that maps sequences in $(\mathbb{R}^{d})^n$ into a finite measurable partitions of $\mathbb{R}^{d}$. The second element is a non-negative threshold denoted by $a_n\in \mathbb{R}^+$. Then given a pair $(\pi_n(\cdot),a_n)$ and some data $z^n_1\in \mathbb{R}^{dn}$,  the data-driven test is constructed as follows: \\
	{\bf 1)} Use the quantization $\pi_n(z^n_1)=\left\{A_i, i=1,..,  \left| \pi_n(z^n_1)\right| \right\} \subset \mathcal{B}(\mathbb{R}^d)$ to estimate $P$ over the cells $\pi_n(z^n_1)$:
			\begin{equation}\label{eq_problem_st_3}
				\hat{P}_n(A) \equiv  \frac{1}{n} \sum_{i=1}^n \mathbf{1}_{A}(z_i) \text { and }
			\end{equation} 
			\begin{equation}\label{eq_problem_st_4}
				\hat{Q}^*_n(\underbrace{A^1\times A^2}_{A=})  \equiv  \hat{P}_n(A^1\times \mathbb{R}^q) \cdot  \hat{P}_n(\mathbb{R}^p \times A^2),
			\end{equation}
		for any $A \in \pi_n(z^n_1)$. In Eq.(\ref{eq_problem_st_4}) we assume that the cells of $\pi_n(z^n_1)$ have a product structure, i.e.,  
		$A=A^1\times A^2$ where $A^1\in \mathcal{B}(\mathbb{R}^p)$ and $A^2\in \mathcal{B}(\mathbb{R}^q)$.\footnote{This event-wise product structure 
		on the cells of $\pi_n(z^n_1)$ is needed to estimate both $P$ and $Q^*$ from i.i.d. samples of $P$. 
		}\\
	{\bf 2)} Project the data $z^n_1$ over the cells of $\pi_n(z^n_1)$. A simple projection (or representation) function is the following: 
		     	\begin{equation}\label{eq_problem_st_5}
				O_{\pi_n}(z) \equiv \sum_{j=1}^{\left| \pi_n(z^n_1)\right|} j \cdot \mathbf{1}_{A_j}(z)  \in  \left\{1,..,\left| \pi_n(z^n_1)\right| \right\}.
			\end{equation}
			The projected data is given and denoted by $o^n_1 \equiv (o_1= O_{\pi_n}(z_1),..,o_n=O_{\pi_n}(z_n))$.\\
	{\bf 3)} Compute the log-likelihood ratio of the empirical distributions in (\ref{eq_problem_st_3}) and (\ref{eq_problem_st_4}) using the quantized 
	 	       (or projected over $\sigma(\pi_n(z^n_1))$) data $o^n_1$.  
		       In particular, we consider the following log-likelihood ratio per sample as  
		     	\begin{equation}\label{eq_problem_st_6}
				\hat{i}_{\pi_n}(o^n_1) \equiv \frac{1}{n} log \frac{\hat{P}_{O_{\pi_n}}(o_1) \hat{P}_{O_{\pi_n}}(o_2) \cdots \hat{P}_{O_{\pi_n}}(o_n)}{\hat{Q}_{O_{\pi_n}}(o_1) \hat{Q}_{O_{\pi_n}}(o_2) \cdots \hat{Q}_{O_{\pi_n}}(o_n)},  
			\end{equation}
			where $\hat{P}_{O_{\pi_n}}$ and $\hat{Q}^*_{O_{\pi_n}}$ denote the empirical distributions of the quantized random variable $O_{\pi_n}$
			in its representation space $\left\{1,..,\left| \pi_n(z^n_1)\right| \right\}$, meaning that by construction $\hat{P}_{O_{\pi_n}}(j) = \hat{P}_n(A_j)$ and
			$\hat{Q}_{O_{\pi_n}}(j) = \hat{Q}^*_n(A_j)$ for all $j\in \left\{1,..,\left| \pi_n(z^n_1)\right| \right\}$.\\
	{\bf 4)}  Finally, the decision is given by the following rule: 
			\begin{equation}\label{eq_problem_st_7}
			\phi_{\pi_n, a_n} (o^n_1) \equiv \left\{
				\begin{array}{lll}
				0 & \textrm{ if } & i_{\pi_n}(o^n_1) < a_n\\
				1 & \textrm{ if } & i_{\pi_n}(o^n_1) \geq a_n
				\end{array}\right..
			\end{equation} 
 The second step introduced $O_{\pi_n}(Z)$ in (\ref{eq_problem_st_5}). This object captures the role played by the data-driven partition ($\pi_n$) in our log-likelihood statistics in (\ref{eq_problem_st_6}).
 %
 
\section{Regret Analysis: The Representation Learning Problem}
\label{sec_regret}
To analyze the quality of the proposed test, in this section we compare the proposed statistics in (\ref{eq_problem_st_6}) 
with the statistics used by an oracle NP test.
We consider the data-driven partition in the pair $(\pi_n(\cdot),a_n)$ as a learning agent. For this analysis, let us first fix a finite partition $\pi= \left\{A_i, i=1,..,\left\{1,..,J \right\} \right\}$. Given i.i.d. realizations $Z_1,..,Z_n$ from $P\in \mathcal{P}(\mathbb{R}^d)$,   
we consider the empirical-quantized log-likelihood statistics in (\ref{eq_problem_st_6}). 
Adopting the notion of regret from universal source coding \cite{csiszar_2004}, 
let us consider as a reference the true likelihood ratio of $P$ with respect to $Q^*(P)$ associated with the expression in (\ref{eq_problem_st_6}) but with no quantization effects, i.e., \footnote{By construction $P\ll Q^*$;  therefore, 
the RN derivative $\frac{dP}{dQ^*}$ is well defined in (\ref{eq_problem_st_8}) where it is clear that if 
$(\log \frac{dP}{dQ^*}(z))_{z\in \mathbb{R}^d} \in \ell_1(P)$, then $I(X,Y)=\mathbb{E}_{Z\sim P}  \left\{ \log \frac{dP}{dQ^*}(Z)) \right\} < \infty$.}
\begin{equation}\label{eq_problem_st_8}
	i_n(z_1,...,z_n) \equiv \frac{1}{n} \log \frac{d P}{d Q^*}(z_1)\cdot \frac{dP}{dQ^*}(z_2) \cdots \frac{dP}{dQ^*}(z_n).
\end{equation} 
$i_n(z_1,...,z_n)$ is the ideal (oracle) information term used by the NP test against independence \cite{csiszar_2004}. 
We measure the (sample-wise) regret of $\pi$ as
\begin{align}\label{eq_problem_st_9}
	i_n(z_1,...,z_n) - \hat{i}_{\pi}(O_{\pi}(z_1),....,O_{\pi}(z_n)), 
\end{align} 
which measures the discrepancy between the empirical-quantized statistics in (\ref{eq_problem_st_6})
and the oracle term in (\ref{eq_problem_st_8}). 

\subsection{Connection with Divergence Estimation}
The regret in (\ref{eq_problem_st_9}) has two error sources:  one associated with the quantization of the space (the representation quality of $\pi$) and the other associated with the fact that we use the empirical distribution $\hat{P}_n$ in (\ref{eq_problem_st_3}) instead of the true model $P$. To isolate these components, it is useful to introduce {\em the oracle-quantized information} given by
\begin{equation}\label{eq_problem_st_10}
	i_{\pi}(o^n_1) \equiv \frac{1}{n} \log \prod_{i=1}^n  \frac{P_{O_{\pi}}(o_i)}{Q^*_{O_{\pi}}(o_i)} \text{ with}\ o^n_1\in  \left\{1,..,J \right\}^n,
\end{equation}
where $P_{O_{\pi}}(j) = P(A_j)$ and $Q^*_{O_{\pi}}(j) = {Q}^*(A_j)$ for all $j\in \left\{1,..,J \right\}$ are short-hand for the true probabilities 
induced by $O_\pi(\cdot)$,  $P$ and $Q^*$ in $\left\{1,..,J \right\}$. Then, the regret in  (\ref{eq_problem_st_9}) can be decomposed in two components:
\begin{align}\label{eq_problem_st_11}
	&i_n(z_1,...,z_n) - \hat{i}_{\pi}(O_{\pi}(z_1),....,O_{\pi}(z_n)) =\nonumber\\ 
	&\underbrace{i_n(z_1,...,z_n)  - i_{\pi}(O_{\pi}(z_1),....,O_{\pi}(z_n))}_{I} + \nonumber\\
	& \underbrace{i_{\pi}(O_{\pi}(z_1),....,O_{\pi}(z_n)) - \hat{i}_{\pi}(O_{\pi}(z_1),....,O_{\pi}(z_n))}_{II}.
\end{align} 
The first term (I) on the right hand side (RHS) of Eq.(\ref{eq_problem_st_11}) captures an information loss attributed to the quantization 
(\ref{eq_problem_st_5}) (the approximation error). This expression convergences as $n$ tends to infinity  to 
$D(P||Q^*) -D_{\sigma(\pi)}(P||Q^*) \geq 0$
almost surely. 
The second term (II) on the RHS of (\ref{eq_problem_st_11}) captures the discrepancy in the information density attributed to the use of empirical distributions.  
Crucially, these two error sources are driven by the partition $\pi$.
At this point, it is worth noticing that the empirical information term can be expressed as
$\hat{i}_{\pi}\left(O_{\pi}(z_1),....,O_{\pi}(z_n)\right) =\sum_{A\in \pi}  \hat{P}_n(A) \log \frac{\hat{P}_n(A)}{\hat{Q}_n^*(A)}$,  
which is a histogram-based estimator of the divergence restricted over the cells of $\pi$ \cite{silva_2012,silva_isit_2007,silva_2010}. 
From these results, the RHS of  (\ref{eq_problem_st_11}) can be re-structured as follows
\begin{align}\label{eq_problem_st_12}
	&i_n(Z_1,...,Z_n) - \hat{i}_{\pi}(O_{\pi}(Z_1),....,O_{\pi}(Z_n)) 	=\nonumber\\ 
	& \underbrace{i_n(Z_1,...,Z_n) - D(P||Q^*)}_{I}  + 
	 \underbrace{D(P||Q^*) -  D_{\sigma({\pi})}({P}||{Q}^*)}_{II} +  \nonumber\\
	& \underbrace{ D_{\sigma({\pi})}({P}||{Q}^*)  - D_{\sigma({\pi})}(\hat{P}_n||\hat{Q}_n^*)}_{III} .
\end{align} 
The first term ($I$) on the RHS of (\ref{eq_problem_st_12})
goes to zero with probability one (by the strong law of large numbers assuming that $D(P||Q^*)<\infty$)  independent of $\pi$. Therefore, from the perspective of evaluating the effect of $\pi$ in the regret, this term can be overlooked. On the other hand, the second term ($II$) captures the approximation error, or what we lost in discrimination (as the  number of samples goes to infinity) when projecting the data over $\sigma(\pi)$ with respect to the lossless  term in (\ref{eq_problem_st_8}). Finally, the last term ($III$) measures the estimation error, which is the discrepancy between the true (oracle) distributions and the empirical distributions in the information divergence sense over events of $\sigma(\pi)$  (see Eq.(\ref{eq_pre_1})). 

\subsubsection{The Representation Problem}
\label{main_representation_problem}
Here we introduce the main design problem of this work, which is to find a data-driven partition that offers an optimal balance between the two relevant terms  presented in (\ref{eq_problem_st_12}). For the estimation error, the idea is to adopt  distribution-free (universal) error bounds for
$\left| D_{\sigma({\pi})}({P}||{Q}^*)  - D_{\sigma({\pi})}(\hat{P}||\hat{Q}^*) \right|$ of the form (more details in Section \ref{sub_sec_reim}): 
\begin{equation*} 
\mathbb{P}  \left( \left| D_{\sigma({\pi})}({P}||{Q}^*)  - D_{\sigma({\pi})}(\hat{P}_n||\hat{Q}_n^*) \right| \leq r_n(\pi)  \right) \geq 1 - \delta,
\end{equation*}
where $r_n(\pi)$ is the confidence interval for the confidence probability $1-\delta$.
Equipped with this result,  we could use the following upper bound for the regret: 
	$D(P||Q^*) - \hat{i}_{\pi}(O_{\pi}(z_1),....,O_{\pi}(z_n))
	\leq   \left[ D(P||Q^*) -  D_{\sigma({\pi})}({P}||{Q}^*)  \right] + r_n(\pi)$, 
with high probability. 
Finally, we could formulate the problem of selecting $\pi$ over a  family of representations $\Pi$ (subset of $\mathcal{Q}(\mathbb{R}^d)$) as
the solution of the following {\em regularized info-max problem}: 
\begin{align}\label{eq_problem_st_13}
	\pi^*_n \equiv \arg\max_{\pi\in \Pi}  D_{\sigma({\pi})}({P}||{Q}^*) - r_n(\pi).
\end{align}

The  learning task in (\ref{eq_problem_st_13}) is still intractable.  It resembles an oracle learner agent (teacher) that selects $\pi$ in $\Pi$ solving the trade-off between the two errors, one of which needs the true model  $P$. 
Section \ref{sec_tsp} addresses a tractable info-max version of 
(\ref{eq_problem_st_13}), where instead of $P$  the empirical distribution $\hat{P}$  in (\ref{eq_problem_st_3}) is adopted.  

\subsection{Strong Consistency: A Basic Requirement}
\label{sub_sec_regret_consistency}
In this subsection, we introduce the idea of measuring the discrepancy between $\hat{i}_{\pi_n}(O_{\pi_n}(z_1),....,O_{\pi_n}(z_n))$ and the ideal (oracle) information $i_n(z_1,...,z_n)$. From this perspective, we could introduce a new notion of consistency based on this learning objective.
\begin{definition}
	\label{def_consistency_suff_stat}
	Our 
	scheme $\left\{(\pi_n, a_n), n\geq 1 \right\}$ 
	is said to be {\em strongly consistent on the regret} 
	if for any $P\in \mathcal{P}(\mathbb{R}^d)$ and i.i.d. process $(Z_n)_{n\geq 1}$ with $Z_i\sim P$, it follows that
	\begin{equation*} 
	\lim_{n \rightarrow \infty}  \left| i_n(Z_1,...,Z_n) - \hat{i}_{\pi_n}(O_{\pi_n}(Z_1),....,O_{\pi_n}(Z_n)) \right| = 0, 
	\end{equation*}
	$\mathbb{P}$-almost surely (a.s.). 
\end{definition}

A simple result follows: 
\begin{proposition}\label{pro_consistency_regret_MI_estimation}
If $\left\{(\pi_n, a_n), n\geq 1 \right\}$  is  {strongly consistent on the regret},  then  $\hat{i}_{\pi_n}(O_{\pi_n}(Z_1),....,O_{\pi_n}(Z_n))$ is a strongly consistent estimator of the MI between $X$ and $Y$.\footnote{From Definition \ref{def_consistency_suff_stat}, it follows directly that $\lim_{n \rightarrow \infty} i_n(Z_1,...,Z_n)  = \lim_{n \rightarrow \infty} D_{\sigma({\pi})}(\hat{P}_n||\hat{Q}_n^*) = D(P||Q^*) = I(X,Y)$, $\mathbb{P}$-almost surely.} 
\end{proposition}

Returning to our original problem, the next result offers sufficient conditions for a data-driven scheme $\left\{ (\pi_n,a_n) n \geq 1 \right\}$ to be strongly consistent for detecting independence (Def. \ref{def_consistency}).

\begin{theorem}\label{th_main}
	Let $\left\{ (\pi_n,a_n), n \geq 1 \right\}$ be the data-driven scheme of Section \ref{sec_plug_in_detector}.  
	 \begin{itemize}
	 	\item[i)] If $(a_n)_n$ is $o(1)$, 
		\item[ii)] $ \left\{ \pi_n, n\geq 1 \right\}$ is strongly consistent on the regret (Def. \ref{def_consistency_suff_stat}),  
		\item[iii)] under $\mathcal{H}_0$,  
		$( \hat{i}_{\pi_n}(O_{\pi_n}(Z_1),....,O_{\pi_n}(Z_n)))_{n\geq 1}$ is $o(a_n)$  in the sense that
		$\lim_{n \rightarrow \infty} \frac{ \hat{i}_{\pi_n}(O_{\pi_n}(Z_1),....,O_{\pi_n}(Z_n))}{a_n}=0$ $\mathbb{P}$-a.s.,
	\end{itemize}
	then $(\phi_{\pi_n,a_n}(O_{\pi_n}(\cdot)))_{ n \geq 1}$ in (\ref{eq_problem_st_7})  is strongly consistent for detecting independence (Def. \ref{def_consistency}).
\end{theorem}
The proof is presented in Appendix \ref{proof_th_main}. 

Consistency is a basic asymptotic requirement that is non-sufficient when looking into practical applications that operate with a finite sample size. Hence, in the next sections, we present a test that is strongly consistent (Th.\ref{th_test_consistency_2}) but also satisfies some relevant non-asymptotic properties in the form of finite-length performance results (Ths. \ref{th_nearly_minimax_opt_regret}, \ref{th_bound_on_tail_of_T0} and Lemma \ref{th_structural_detection_ind}).

\section{Tree-Structured Data-Driven Partitions}
\label{sec_tsp}
In this section, an empirical version of  (\ref{eq_problem_st_13}) is studied considering for $\Pi$ a dictionary of tree-structured data-driven partitions (TSP). This TSP family was introduced in \cite{silva_2012} for the problem of MI estimation.
The focus of this section is to demonstrate its  potential for the problem of testing independence. 

\subsection{The Collection of Partitions}
\label{sub_sec_tsp_grow}
The construction of this family begins by defining a ``root" node to index the trivial partition $A_{root} \equiv \left\{ \mathbb{R}^d \right\}$.  The process continues by selecting a coordinate in $\left\{1,..,d \right\}$ to project the data (1D projection), then order the projected data (scalar values), select the median of the ordered sequence, and finally create a statistically equivalent partition of $A_{root}$ using the selected coordinate axis and the median to split the cell. Two new cells are created from $A_{root}$ with almost half the sample points in each (exactly half when $n$ is an even integer). 

These new cells are indexed as the left and right children of the ``root" denoted by $(l(root),r(root))$; i.e., the new cells are $A_{l(root)}$ and $A_{r(root)}$ where $A_{root}=A_{l(root)} \cup A_{r(root)}$.  The process continues by selecting a new coordinate in $\left\{1,..,d \right\}$, where the statistically equivalent  binary splitting criterion is iterated in $A_{l(root)}$ and $A_{r(root)}$ until a stopping condition is met. The stopping criterion adopted here  imposed a minimum empirical probability in each created cell that we denote by $b_n \in (0,1)$.  Therefore, at the end of this growing binary process, a collection of nodes is produced $\mathcal{I} \equiv \left\{root, l(root),r(root), ...\right\}$ associated with a binary tree (by construction) and  the nodes'  respective cells $\left\{A_v, v\in \mathcal{I} \right\}$ where $\hat{P}_n(A_v)\geq b_n$ for any $v\in \mathcal{I}$. 

Using the Breiman {\em et al.} \cite{breiman_1984} convention,  a binary tree $T$ is a collection of nodes in $\mathcal{I}$: one node of degree 2 (the "root"), and the remaining nodes of degree 3 (internal nodes) or degree 1 (leaf or terminal nodes) \cite{breiman_1984}. In this convention, the full-tree is denoted and given by $T^{full}_{b_n} \equiv \mathcal{I}$. Importantly, if $\tilde{T}\subset T^{full}_{b_n}$ and  $\tilde{T}$ is a binary tree by itself, then we say that $\tilde{T}$ is a subtree of $T^{full}_{b_n}$. Moreover, if both have the same root, we say that $\tilde{T}$ is a \textit{pruned version} of $T^{full}_{b_n}$, and we denote this by $\tilde{T} \ll T^{full}_{b_n}$.  Finally, if we denote by $\mathcal{L}(T)$ the leaf nodes of an arbitrary tree $T \ll T^{full}_{b_n}$,  it is simple to verify that  
$\pi_T\equiv \left\{A_v, v\in \mathcal{L}(T) \right\}$ is a data-driven partition of $\mathbb{R}^d$ indexed by $T$ where every cell in $\pi_T$  has the desired product structure in Eq.(\ref{eq_problem_st_4}) \cite{silva_2012}.\footnote{The interested reader is referenced to \cite{silva_2012} for a systematic explanation of this TSP construction.} 

\subsection{Regularized (Empirical) Information Maximization}
\label{sub_sec_reim}
To address the info-max design objective in (\ref{eq_problem_st_13}) 
using our (data-driven) tree-indexed family $\Pi_n \equiv \left\{  \pi_T:  T\ll T^{full}_{b_n}\right\}$,  we have to first find an expression for $r_n(\pi)$ in  (\ref{eq_problem_st_13}) for any $\pi_T \in \Pi_n$.  The next result offers the following:
\begin{lemma}\label{th_est_err_bound} (Silva {\em et al.}\cite[Th. 1]{silva_2012})
Let $\mathcal{G}^k_{b_n} \equiv \left\{T \ll T^{full}_{b_n}: \left|T\right|=k \right\}$ be
the family of pruned TSPs of size $k$. Then, $\forall k \in \left\{1,..,\left|T^{full}_{b_n}\right|\right\}$,  $\forall n >0$, 
and any small $\delta \in (0,1)$, there is a threshold $\epsilon_c(\cdot)$ where 
\begin{align}\label{eq_sub_sec_reim_1}
	&\mathbb{P}\left(\sup_{T \in \mathcal{G}^k_{b_n}} \left|{ D_{\sigma({\pi_T})}({P}||{Q}^*)  - D_{\sigma({\pi_T})}(\hat{P}_n||\hat{Q}_n^*) }\right| \leq  \epsilon_c(\cdot) \right) \geq \nonumber\\
	&1-\delta,
\end{align}
where specifically $\epsilon_c(n,b_n, d, \delta,k) =$
\begin{align}\label{eq_sub_sec_reim_2}
\frac{24 \sqrt{2}}{b_n\sqrt{n}} \sqrt{ \ln(8/\delta) + k  \left[(d+1) \ln(2)+d \ln(n) \right]} .
\end{align}
\end{lemma}
Importantly the bound in (\ref{eq_sub_sec_reim_1}) is distribution-free \cite{silva_2012},
and a function of $b_n$ and $k$ (the size of the family $\mathcal{G}^k_{b_n}$).  
From this distribution-free concentration result, the union bound tells us that the following events in $\mathbb{R}^{dn}$
\begin{align}\label{eq_sub_sec_reim_3}
 \left\{
 \sup_{T \in \mathcal{G}^k_{b_n}} \left| D_{\sigma({\pi_T})}({P}||{Q}^*)  - D_{\sigma({\pi_T})}(\hat{P}_n||\hat{Q}_n^*) \right| \leq  r_{b_n, \delta}(k)\right\},  
\end{align}
with $r_{b_n, \delta}(k) \equiv  \epsilon_c(n,b_n, d, \delta \cdot b_n, k)$,  happen simultaneously for any $k=1,..,\left| T^{full}_{b_n} \right|$ with  probability at least $1-\delta$ (with respect to $\mathbb{P}$). Equipped with these penalizations, i.e., $r_{b_n, \delta}(\left|  T \right| )$ in (\ref{eq_sub_sec_reim_3}) for any $T \ll T^{full}_{b_n}$,  the empirical version of the info-max problem in (\ref{eq_problem_st_13}) 
is
\begin{align}\label{eq_sub_sec_reim_4}
	\hat{T}_{b_n,\delta_n} \equiv \arg\max_{T \ll T^{full}_{b_n}}  D_{\sigma({\pi_T})}(\hat{P}_n||\hat{Q}_n^*) - r_{b_n, \delta_n}(\left| T \right|).
\end{align}

Finally, it is important to note that both $(b_n)_{n\geq 1}$ and $(\delta_n)_{n\geq 1}$ determine the trees $(\hat{T}_{b_n,\delta_n})_{n\geq 1}$  and  TS partitions that we denote here by $({\pi}_{b_n,\delta_n})_{n\geq 1}$. In addition,  if we include the thresholds $(a_n)_{n\geq 1}$, we denote by $(\phi_{b_n,\delta_n,a_n}(\cdot))_{n\geq 1}$ the rules induced by $({\pi}_{b_n,\delta_n})_{n\geq 1}$ and $(a_n)_{n\geq 1}$ in  (\ref{eq_problem_st_7}). 

\subsection{Consistency: A Preliminary Analysis}
\label{sec_consistency}
The following results show that  the TSP scheme in (\ref{eq_sub_sec_reim_4}) is strongly consistent on the regret (Def.\ref{def_consistency_suff_stat}).  
\begin{lemma}\label{th_consistency_regret}
Let us assume that $P$ has a density\footnote{$P$ is absolutely continuous with respect to the Lebesgue measure in $(\mathbb{R}^d, \mathcal{B}(\mathbb{R}^d))$.} in $\mathbb{R}^d$. \\
{\bf i)} Under the conditions that $(b_n) \approx (n^{-l})$ for $l \in (0, {1}/{3})$,	$(\delta_n)$ is $o(1)$ and $(1/\delta_n)$ is $\mathcal{O}({e}^{n^{1/3}})$, we have that
	\begin{align}\label{eq_main_results_2}
		&\lim_{n \rightarrow \infty}\left| i_n(Z_1,...,Z_n) - \hat{i}_{\pi_{b_n,\delta_n}}(O_{\pi_{b_n,\delta_n}}(Z_1),....,O_{\pi_{b_n,\delta_n}}(Z_n)) \right| \nonumber\\
		&=0,\ \mathbb{P}\text{-a.s.}.
	\end{align}
{\bf ii)} Assuming that $X$ and $Y$ are independent ($\mathcal{H}_0$), if we select $(1/\delta_n)_n \approx (e^{n^{1/3}})$,  
it follows that for any $p>0$ 
	\begin{align}\label{eq_main_results_3}
		  (\hat{i}_{\pi_{b_n,\delta_n}}(O_{\pi_{b_n,\delta_n}}(Z_1),....,O_{\pi_{b_n,\delta_n}}(Z_n)))_{n\geq 1}  \text{ is } o(n^{-p}), \mathbb{P}\text{-a.s.}.
	\end{align}	
\end{lemma}
The proof is presented in Appendix \ref{proof_th_consistency_regret}.

In addition to achieve consistency on the regret, stated in part i), the part ii) of this result shows that under $\mathcal{H}_0$ the regret achieves a super polynomial velocity of convergence to zero: $i_n(Z_1,...,Z_n) =0$ with probability one under $\mathcal{H}_0$.  Therefore, 
the empirical information rate $\hat{i}_{\pi_{b_n,\delta_n}}(O_{\pi_{b_n,\delta_n}}(Z_1),....,O_{\pi_{b_n,\delta_n}}(Z_n))$ tends to zero faster than any polynomial order in $n$ with probability one, which is a remarkable capacity to detect this condition from data.
In light of Theorem \ref{th_main},  we could have a range of admissible vanishing thresholds $(a_n)_{n\geq 1}$ where strong consistency for detecting independence can be achieved (Def. \ref{def_consistency}). 
This is stated in the following result. 
\begin{corollary}\label{th_test_consistency}
Let us assume the setting and conditions on $(b_n)_{n\geq 1}$ and $(\delta_n)_{n\geq 1}$ stated in Lemma \ref{th_consistency_regret} part ii) for the TSP scheme.
If $(a_n)_{n\geq 1}$ 
is $\mathcal{O}(n^{-q})$ for some arbitrary  $q>0$, then the TSP scheme $\left\{ \phi_{{b_n,\delta_n}, a_n}, n\geq 1 \right\} $ is strongly consistent for detecting independence (Def. \ref{def_consistency}). 
\end{corollary}
\begin{proof} The proof derives directly from Theorem \ref{th_main} and the two results obtained in Lemma \ref{th_consistency_regret}.
\end{proof}

\section{Main Finite-Length Results}
\label{sec_finite_length}
In this section, we focus on finite-length (non-asymptotic) results. We establish conditions where the solution of (\ref{eq_sub_sec_reim_4}) nearly matches the performance of its equivalent oracle version in (\ref{eq_problem_st_13}) with high probability. This non-asymptotic result is instrumental to show later one of the main findings of our work: the capacity of our scheme to detect $\mathcal{H}_0$ with the structure of the partition.

\begin{theorem}\label{th_nearly_minimax_opt_regret}
	Under  the conditions that
	$(b_n) \approx (n^{-l})$ for $l \in (0, {1}/{3})$,  
	$(\delta_n)$ is $o(1)$ and $(1/\delta_n)$ is $\mathcal{O}({e}^{n^{1/3}})$,
	we have that\\ 
	{\bf i)} under $\mathcal{H}_0$:  for any $\epsilon>0$ there is $N(\epsilon)>0$ such that $\forall n \geq N(\epsilon)$, the equality
	\begin{align}\label{eq_main_results_0}
	 	\underbrace{\hat{i}_{\pi_{b_n,\delta_n}}(O_{\pi_{b_n,\delta_n}}(Z_1),....,O_{\pi_{b_n,\delta_n}}(Z_n))=i_n(Z_1,...,Z_n)}_{\text{zero regret regime}}=0,   
	\end{align}	 
	holds with $\mathbb{P}$-probability $1-\epsilon$. 
	
	{\bf ii)}
	under $\mathcal{H}_1$: for any $\epsilon>0$ there is  $N(\epsilon)$ such that $\forall n \geq N(\epsilon)$ the bound
	\begin{align}\label{eq_main_results_1}
		& i_n(Z_1,...,Z_n) - \hat{i}_{\pi_{b_n,\delta_n}}(O_{\pi_{b_n,\delta_n}}(Z_1),....,O_{\pi_{b_n,\delta_n}}(Z_n) )\nonumber\\
		&\leq   i_n(Z_1,...,Z_n) - I(X,Y)  + \nonumber\\
		&\min_{T\ll T^{full}_{b_n}} \left[ D(P||Q^*(P)) -  D_{\sigma({\pi_T})}({P}||{Q}^*(P))  \right] + 2 r_{b_n, \delta_n}(  \left| T \right| ).
	\end{align}
	holds with $\mathbb{P}$-probability $1-\epsilon$.
\end{theorem}
The proof of this result is presented in Appendix \ref{proof_th_nearly_minimax_opt_regret}.

This result presents two optimality bounds for the information term of the TSP scheme under the two main hypotheses 
($\mathcal{H}_0$ and $\mathcal{H}_1$) of our problem. Under $\mathcal{H}_0$, our regularization approach is capable of detecting this structure (with an arbitrary high probability $1-\epsilon$) in the sense that  $\hat{T}_{b_n,\delta_n}$ in (\ref{eq_sub_sec_reim_4}) reduces to the trivial partition $\left\{\mathbb{R}^d \right\}$.  From this, we obtain the  zero regret condition $\hat{i}_{\pi_{b_n,\delta_n}}(O_{\pi_{b_n,\delta_n}}(Z_1),....,O_{\pi_{b_n,\delta_n}}(Z_n))=i_n(Z_1,...,Z_n)$.  This means that the solution of $\hat{T}_{b_n,\delta_n}$ by itself (with no threshold) is capable of detecting independence. 

On the other hand, under $\mathcal{H}_1$, we notice that it is only relevant  to bound the under-estimation of $i_n(Z_1,...,Z_n)$ with the empirical (log-likelihood ratio) information $ \hat{i}_{\pi_{b_n,\delta_n}}(O_{\pi_{b_n,\delta_n}}(Z_1),....,O_{\pi_{b_n,\delta_n}}(Z_n)) $. Limiting the analysis only to the underestimation of $i_n(Z_1,...,Z_n)$ can be argued from the well-known observation that quantization (on average) provides a bias  (under-estimation) estimation of $\mathbb{E}(i_n(Z_1,...,Z_n))=I(X;Y)$.  More importantly for our problem, an overestimation of the oracle information $i_n(Z_1,...,Z_n)$ does not  increase the type 2 error when using $\hat{i}_{\pi_{b_n,\delta_n}}(O_{\pi_{b_n,\delta_n}}(Z_1),....,O_{\pi_{b_n,\delta_n}}(Z_n))$ instead of the oracle $i_n(Z_1,...,Z_n)$ under $\mathcal{H}_1$. 

In summary, Theorem \ref{th_nearly_minimax_opt_regret} indicates that with an arbitrary high probability, the empirical TSP  solution of (\ref{eq_sub_sec_reim_4}) achieves the performance of the oracle solution that optimizes the upper bound on the regret derived  in Eq.(\ref{eq_problem_st_13}) over the TSP family 
$\left\{\pi_T, T\ll T^{full}_{b_n} \right\}$.

\begin{remark}
In the proof of Theorem \ref{th_nearly_minimax_opt_regret}, we determine a set $\mathcal{E}^n_{\delta_n, b_n}$ (see  Eq.(\ref{eq_th_nearly_minimax_opt_regret_2})) where if 
$z^n_1\in \mathcal{E}^n_{\delta_n, b_n}$ then (\ref{eq_main_results_0}) holds under $\mathcal{H}_0$ and  (\ref{eq_main_results_1}) holds under $\mathcal{H}_1$. Importantly, we obtain that  $\mathbb{P}(\mathcal{E}^n_{\delta_n, b_n})\geq 1 -\delta_n$ from Lemma \ref{th_est_err_bound}. Therefore, in Theorem \ref{th_nearly_minimax_opt_regret}, $N(\epsilon)$ is fully determined by $(\delta_n)_{n\geq 1}$ for any $\epsilon>0$. Indeed, $N(\epsilon)= \inf  \left\{n\geq 1, st.\  \delta_n<\epsilon \right\}$, which is well-defined since $(\delta_n)$ is $o(1)$. In particular,  for the case $(1/\delta_n)$ being $\mathcal{O}(e^{n^{1/3}})$, it follows that $N(\epsilon)$ is $\mathcal{O}(\ln (1/\epsilon)^3)$.
\end{remark}

\subsection{Detecting Independence with the Structure of $\hat{T}_{b_n,\delta_n}$}
\label{sub_sec_structural_detection_of_H0}
From  (\ref{eq_main_results_0}), our data-driven partition 
has the capacity to detect independence (under $\mathcal{H}_0$) in a stronger structural way: 
(with high probability) the data-driven partition collapses to the trivial cell, i.e.,  $\pi_{b_n,\delta_n}=\left\{\mathbb{R}^d \right\}$, with a finite sample size.  Here, we improve this result by showing 
that the condition $\pi_{b_n,\delta_n}=\left\{\mathbb{R}^d \right\}$ happens within a finite sample size almost surely.

Let us introduce the (random) time at which the partition process $(\pi_{b_n,\delta_n})_{n\geq 1}$ collapses to the trivial set  $A_{root}=\left\{\mathbb{R}^d \right\}$: 
\begin{definition} $\mathcal{T}_0((Z_n)_{n\geq 1})  \equiv$
\begin{equation} \label{eq_sub_sec_structural_detection_of_H0_1}
	 \sup {\left\{ m\geq 1:    \left| \hat{T}_{b_m,\delta_m} \right| > 1 \right\}} \in \mathbb{N}^* \equiv \mathbb{N}\cup  \left\{\infty \right\}.
\end{equation}
\end{definition}
If $\mathcal{T}_0((Z_n)_{n\geq 1})=k$,  it means that $\left| \hat{T}_{b_n,\delta_n} \right| =1 \Leftrightarrow \pi_{b_n,\delta_n}=\left\{\mathbb{R}^d \right\}$ for any $n > k$.  Therefore, if  $\mathcal{T}_0((Z_n)_{n\geq 1}) \in \mathbb{N}$, this value is expressing the last time the data-driven partition is different from the trivial solution $\left\{\mathbb{R}^d \right\}$. On the other hand, if  $\mathcal{T}_0((Z_n)_{n\geq 1})=\infty$, this condition means that  non-trivial partitions are observed infinitely often (i.o.) in  the sequence $(\pi_{b_n,\delta_n}(Z^n))_{n\geq 1}$. In general, we could have that  $\mathbb{P}(\mathcal{T}_0((Z_n)_{n\geq 1})=\infty)>0$ under $\mathcal{H}_0$, which does not contradict the capacity  that $(\phi_{b_n,\delta_n, a_n}(\cdot))_{n\geq 1}$ has for detecting independence consistently under $\mathcal{H}_0$ (see Corollary \ref{th_test_consistency}).

The following result shows that $\mathcal{T}_0((Z_n)_{n\geq 1}) $ is finite with probability one under some mild conditions. 
\begin{lemma}\label{th_structural_detection_ind}
	Let us assume that $(b_n) \approx (n^{-l})$ for $l \in (0, {1}/{3})$ and
	$(\delta_n)$ is $\ell_1(\mathbb{N})$,  $(1/\delta_n)$ is $\mathcal{O}({e}^{n^{1/3}})$. Then under the hypothesis that $P\in \mathcal{P}_0$: 
		$\mathbb{P}(\mathcal{T}_0((Z_n)_{n\geq 1})<\infty)=1$.
\end{lemma}
The proof is presented in Appendix \ref{proof_th_structural_detection_ind}.

Importantly, the condition $\mathbb{P}(\mathcal{T}_0((Z_n)_{n\geq 1})<\infty)=1$ implies that $(\phi_{{b_n,\delta_n}, a_n} (Z_1,....,Z_n))_{n\geq 1}$ reaches $0$ eventually with probability one for any sequence $(a_n)_{n\geq 1}\in [0,1]^{\mathbb{N}}$. From this observation, we obtain the following result that  improves the regime of parameters where a consistent test is achieved (established in Corollary \ref{th_test_consistency}).
\begin{theorem}\label{th_test_consistency_2}
If
	$(b_n)_{n\geq 1} \approx (n^{-l})$ for $l \in (0, {1}/{3})$, 
	$(\delta_n)_{n\geq 1}$ is $\ell_1(\mathbb{N})$,  $(1/\delta_n)_{n\geq 1}$ is $\mathcal{O}({e}^{n^{1/3}})$, and 
	$(a_n)_{n\geq 1}$ is $o(1)$, 
then the scheme $(\phi_{{b_n,\delta_n}, a_n} (Z_1,....,Z_n))_{n\geq 1}$ is strongly consistent for detecting independence (Def. \ref{def_consistency}).
\end{theorem}
The proof is presented in Appendix \ref{proof_th_test_consistency_2}.

Finally, if we focus on the admissible solution where $(1/\delta_n)\approx (e^{n^{1/3}})$, we have the following refined description for the 
distribution of $\mathcal{T}_0((Z_n)_{n\geq 1})$ under $\mathcal{H}_0$.  
\begin{theorem}\label{th_bound_on_tail_of_T0}
Under the assumption of Theorem \ref{th_structural_detection_ind}, let us consider  that $(1/\delta_n)\approx (e^{n^{1/3}})$.  Under $\mathcal{H}_0$, we have that  $\mathbb{P}(\mathcal{T}_0((Z_n)_{n\geq 1})\geq m) \leq K e^{-m^{1/3}} \sim \mathcal{O}(e^{-m^{1/3}})$ for any $m \geq 1$ and  for some universal constant  $K>0$. 
\end{theorem}
The proof is presented in Appendix \ref{proof_th_bound_on_tail_of_T0}.

\subsection{Remarks about Theorem \ref{th_bound_on_tail_of_T0}:}
{\bf 1:} If we introduce $\mathcal{T}((Z_n)_{n\geq 1}, (a_n)_{n\geq 1})  \equiv$
\begin{equation}\label{eq_sub_sec_structural_detection_of_H0_2}
	 \sup {\left\{ m\geq 1:  \phi_{b_m,\delta_m,a_m}(Z_1,..,Z_m))= 1 \right\}},
\end{equation}
as the first time when the binary process $(\phi_{b_n, \delta_n,a_n}(Z_1,..,Z_n))_{n\geq 1}$ reaches and stays at $0$,  it is simple to  verify that 
$\mathbb{P}(\mathcal{T}((Z_n)_{n\geq 1}, (a_n)_{n\geq 1})\geq m) \leq \mathbb{P}(\mathcal{T}_0((Z_n)_{n\geq 1})\geq m)$
for any $(a_n)_n$ (see Appendix \ref{proof_th_test_consistency_2}). Therefore, under $\mathcal{H}_0$ and the assumptions of Theorem \ref{th_bound_on_tail_of_T0}, we have  as a corollary that $\mathbb{P}(\mathcal{T}((Z_n)_{n\geq 1}, (a_n)_{n\geq 1}) \geq m) \leq K e^{-m^{1/3}} \sim \mathcal{O}(e^{-m^{1/3}})$ for any sequence  $(a_n)_{n\geq 1}$ as long as $(a_n)_{n\geq 1}$ is $o(1)$.\\
{\bf 2:} With this bound, we could determine (under $\mathcal{H}_0$) a lower bound on the number of samples needed to guarantee that we detect independence structurally, i.e., reaching $\pi_{b_n,\delta_n}=\left\{\mathbb{R}^d \right\}$, and also with our schemes $(\phi_{b_n, \delta_n,a_n}(\cdot))_{n\geq 1}$ with a confidence probability $1-\epsilon$. This critical number of samples is achieved for  $m(\epsilon)$ such that 
$K e^{-m(\epsilon)^{1/3}}<\epsilon$ and $K e^{-(m(\epsilon)-1)^{1/3}} \geq \epsilon$. This number scales like $(m(\epsilon))_{\epsilon} \approx \left(\ln(K/\epsilon) \right)^3\sim \mathcal{O}(\ln(1/\epsilon)^3)$.

\section{Implementation}
\label{sec_implementation}

\subsection{The Learning Algorithm}
\label{sub_sec_tsp_algorithm}
Let us briefly revisit the stages presented in Section \ref{sec_tsp} for the construction of our TSP scheme. Beginning with the partition, there are two phases: growing and pruning.  In the growing phase, presented in Section \ref{sub_sec_tsp_grow}, we have a collection of tree-indexed partitions $\left\{\pi_T, T\ll T^{full}_{b_n} \right\}$ where $\pi_T = \left\{A_v, v\in \mathcal{L}(T) \right\}$ and by construction $\hat{P}_n(A_v)\geq b_n = w\cdot n^{-l}$ for any $v\in T$.  Here, we added a scalar parameter $w>0$ as a relevant design element for our numerical analysis. In the pruning phase, we implement the following  complexity-regularized information maximization:
\begin{align}\label{eq_sub_sec_tsp_algorithm_1}
	\hat{T}_{b_n,\delta_n} (\alpha) \equiv \arg\max_{T \ll T^{full}_{b_n}}  D_{\sigma({\pi_T})}(\hat{P}_n||\hat{Q}_n^*) - \alpha \cdot r_{b_n, \delta_n}(\left| T \right|), 
\end{align}
where we included a regularized parameter $\alpha\in \mathbb{R}^+$. 

\begin{figure*}
\centering
\includegraphics[width=1.00\textwidth]{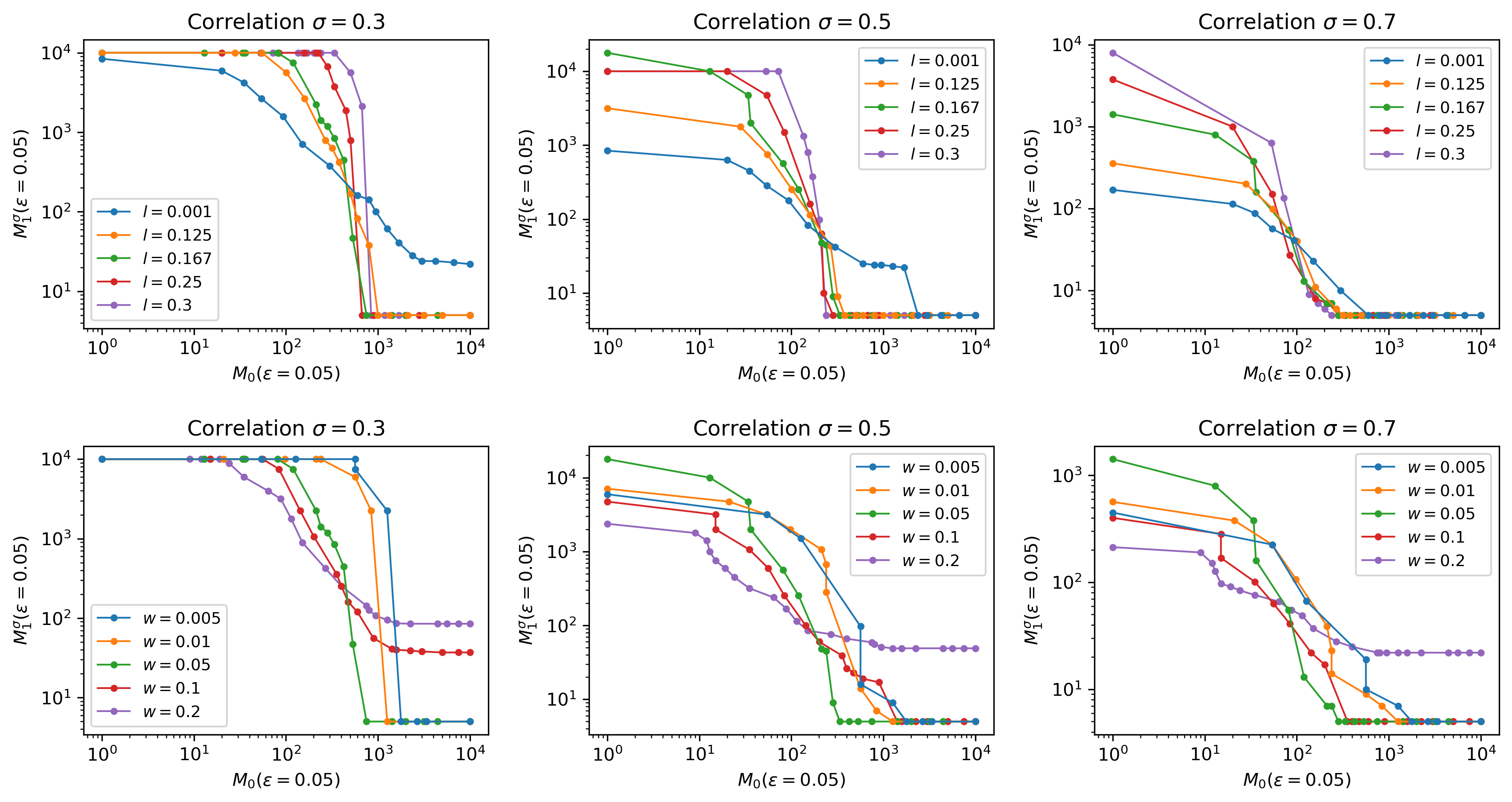}
\caption{Illustration of the trade-off between $M_0(\epsilon=0.05)$ and $M^\sigma_1(\epsilon=0.05)$, presented in Eqs. (\ref{eq_sub_I_test_2}) and (\ref{eq_sub_I_test_2b}),  for different parameters of our TSP approach. Each curve is associated with fixed values of $l$ and $w$ and is produced using different values of $\alpha$.  Three scenarios of correlation (under  $\mathcal{H}_1$) are $\sigma = \mathbb{E}(XY) \in \left\{0.3,0.5,0.7 \right\}$. For the graph in the first row $w=0.05$, whereas $l=0.167$ in the graph depicted in the second row.}
\label{fig4}
\end{figure*}

\subsection{Computational Cost}
To analize the cost of implementing (\ref{eq_sub_sec_tsp_algorithm_1}), 
we analyze the cost of each of its two phases individually. 

\textbf{Growing Phase:} During the growing stage (see Section~\ref{sub_sec_tsp_grow}), we obtain a collection of tree-indexed partitions $\left\{\pi_T, T\ll T^{full}_{b_n} \right\}$, where $\pi_T = \left\{A_v, v\in \mathcal{L}(T) \right\}$. The full tree $T^{full}_{b_n}$ is obtained after a series of statistically equivalent binary partitions. The computational cost of the growing phase depends on the number of sub-partitions (or binary splittings) needed to obtain the full tree (which depends on the number of cells), and the cost of sorting the projected data of each cell and selecting the median of the ordered sequence to partition the cell.
The number of cells in the full tree $\left|\pi_{T^{full}_{b_n}}\right|$ can be estimated considering that our  design criterion states that $\hat{P}_n(A_v)\geq b_n = w\cdot n^{-l}$ for any $v\in T$, where $w \in \mathbb{R}^+$ and $l \in (0, 1/3)$ are design parameters of our method. Then, we should have at least $b_n\cdot n $ samples per cell. Therefore, the number of cells associated with $T^{full}_{b_n}$ is bounded by
\begin{equation}
    \left|\pi_{T^{full}_{b_n}}\right| \leq \frac{n}{b_n\cdot n} = \frac{1}{b_n} \approx n^{l}
\end{equation}
On the other hand, the cost of sorting the projected data of each cell depends on the number of data points to be ordered. In particular, if a cell has $m$ points, then the sorting cost is $\mathcal{O}(m\log(m))$ \cite{cormen_2009}. As we reach deeper levels in the tree, the number of points per cell reduces. If we consider a level of depth $k$ (where $k=1$ corresponds to the root and $k=\left\lceil \log\left(\left|\pi_{T^{full}_{b_n}}\right|\right) \right \rceil \approx \lceil\log(n^l)\rceil$ corresponds to the full tree), there are $2^{k-1}$ cells and each cell contains roughly $n/2^k$ points. Therefore, the order of the total number of operations during the growing phase can be approximated to
\begin{equation*}
    \mathcal{O}\left(\sum_{k=1}^{\lceil\log(n^l)\rceil} 2^{k-1} \cdot \frac{n}{2^{k-1}} \log\left( \frac{n}{2^{k-1}} \right) \right)
    =\mathcal{O}\left(n \log^2(n) \right)
\end{equation*}

 \textbf{Pruning Phase:} The main regularized info-max problem in Eq. (\ref{eq_sub_sec_tsp_algorithm_1}) is solved in this stage.
The first term $D_{\sigma({\pi_T})}(\hat{P}_n||\hat{Q}_n^*) $ is additive (in the sense presented in \cite{scott_2005}), whereas the penalty $r_{b_n, \delta_n}(\left| T \right|)$ scales like $\mathcal{O}(\sqrt{ \left| T \right|})$, and, therefore, it is a sub-additive function of the size of the tree. In this scenario, 
Scott \cite[Th. 1]{scott_2005} showed that the family of minimum cost trees $\{R_i\}_{i=1}^m$: 
\begin{equation}
\label{eq_sub_sec_tsp_algorithm}
R_i \equiv \arg\max_{\substack {T \ll T^{full}_{b_n} \\ |T|=m-i+1}}  D_{\sigma({\pi_T})}(\hat{P}_n||\hat{Q}_n^*) \quad , \, i \in \{1\dots m\}
\end{equation}
is embedded, i.e., $R_1=T^{full}_{b_n} \gg R_2 \gg \ldots \gg R_m= \left\{A_{root}\right\}$. Furthermore, \cite[Th. 2]{scott_2005} states that the solution of our problem in (\ref{eq_sub_sec_tsp_algorithm_1}) corresponds to one of these embedded trees: i.e., for any $\alpha>0$,  $\exists i \in \{1\dots m\} $ such that $\hat{T}_{b_n, \delta_n} (\alpha) = R_i$. This embedded property  allows  computationally efficient algorithms to be designed. Indeed, Scott \cite{scott_2005} presented two algorithms to solve (\ref{eq_sub_sec_tsp_algorithm_1}) with a worst case cost of $\mathcal{O}\left(\left|T^{full}_{b_n}\right|^2\right)$. In our case we have that $\left|T^{full}_{b_n}\right| \leq 1/b_n$, then the pruning stage in (\ref{eq_sub_sec_tsp_algorithm_1}) has a computational cost of $\mathcal{O}\left(n^{2/3}\right)$, which is a polynomial (sub-linear) function of $n$. The pseudo code of this stage is presented in Algorithm \ref{alg:tsp}.

In summary, the construction of our partition (growing and pruning) has a computational cost that is $\mathcal{O}(n\log^2(n))$ on the sample size $n$.

\begin{algorithm} 
\SetAlgoLined
\SetKwInOut{Input}{Input}\SetKwInOut{Output}{Output}
\Input{full tree $T^{full}_{b_n}$, minimum probability of partitions $b_n$, confidence probability $\delta_n$, regularization parameter $\alpha$}\Output{pruned tree $\hat{T}_{b_n,\delta_n}$ in  {\bf Eq.(\ref{eq_sub_sec_tsp_algorithm_1})} 
}
\BlankLine
 Initialization\;
 $\hat{T}_{b_n}^{1}=\{root\}$\;
 Family pruning problem of embedded trees\;
 \For{$k=2$ \KwTo $|T^{full}_{b_n}|$}{
   Maximum empirical information tree of size $k$\;
   $v^*=\underset{v\in\mathcal{L}(\hat{T}_{b_n}^{k-1})}{\argmax} D_{\sigma{_\big(}\pi_{\hat{T}_{b_n}^{k-1} \cup \{l(v),r(v)\}}{_\big)}}(\hat{P}_n||\hat{Q}_n^*)$\;
   $\hat{T}_{b_n}^{k}=\hat{T}_{b_n}^{k-1}\cup \{l(v^*),r(v^*)\}$\;
 }
 Regularized empirical information maximization\;
 $\hat{T}_{b_n,\delta_n}=\underset{T\in\left\{\hat{T}_{b_n}^1,...,\hat{T}_{b_n}^{\left|T_{b_n}^{full}\right|}\right\}}{\argmax} D_{\sigma(\pi_T)}(\hat{P}_n||\hat{Q}_n^*)-\alpha \cdot r_{b_n,\delta_n}(|T|)$\tcp*[r]{$r_{b_n,\delta_n}(|\hat{T}_{b_n}^1|)=0$}
 \caption{TSP family pruning}
 \label{alg:tsp}
\end{algorithm}

\begin{figure*}[ht]
\centering
\includegraphics[width=1.00\textwidth]{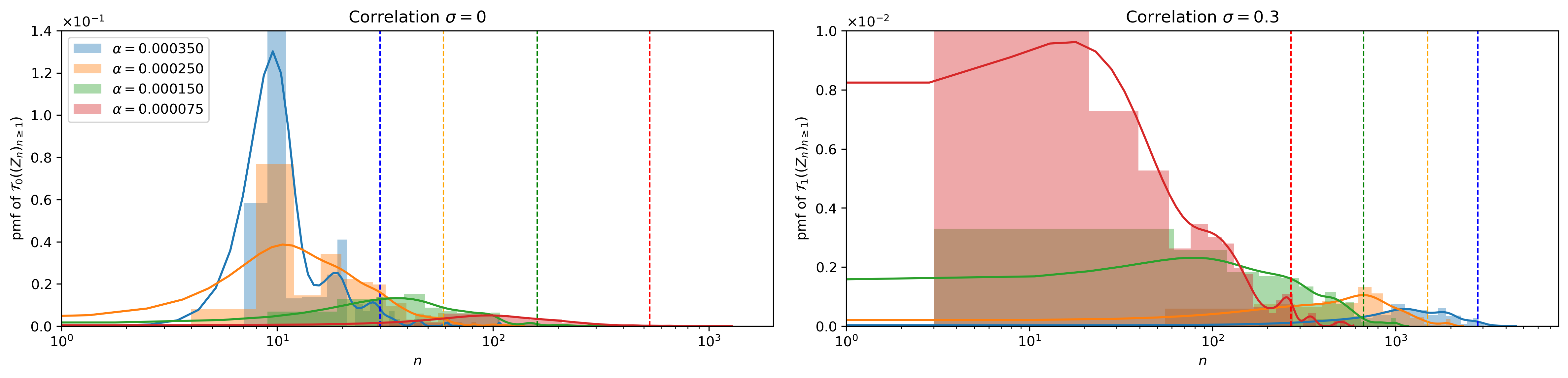}
\caption{Empirical estimation of the probability mass functions of $\mathcal{\tilde{T}}_0((Z_n)_{n\geq 1})$ in Eq.(\ref{eq_sub_I_test_1}) and $\mathcal{\tilde{T}}_1((Z_n)_{n\geq 1})$ in Eq.(\ref{eq_sub_I_test_1b}). Four scenarios of the regularization parameter $\alpha$ are considered to illustrate its effects in the distribution of $\mathcal{\tilde{T}}_0((Z_n)_{n\geq 1})$ and $\mathcal{\tilde{T}}_1((Z_n)_{n\geq 1})$. For each case, the values of $M_0(\epsilon)$ and $M_1^{\sigma}(\epsilon)$ are also illustrated (vertical lines) considering $\epsilon=0.05$ and $\sigma = 0.3$.}
\label{fig5}
\end{figure*}

\section{Empirical Analysis}
\label{sec_implementation_simulations}
In this section, we present some controlled synthetic scenarios to evaluate the performance of our method. We begin analyzing how the selection of parameters affects the capacity of our scheme to estimate MI\footnote{Given the space limit, this preliminary analysis on MI estimation is relegated to the Supplemental Material - Appendix \ref{sub_mi_estimation}.}. From these results and insights, we analyze the performance of the solutions $\left\{ \hat{T}_{b_n,\delta_n} (\alpha), \alpha \geq 0 \right\}$ for testing independence. For the rest of this section, $\pi_{b_n,\delta_n}^\alpha(\cdot)$ denotes the TS partition, $\hat{i}_{b_n,\delta_n}^\alpha(\cdot)$ denotes the MI estimator, and $\phi_{b_n,\delta_n, a_n}^\alpha(\cdot)$ denotes the final test. 

\subsection{Testing Independence: Non-Asymptotic Empirical Analysis}
\label{sub_I_test}
We analyze the problem of testing independence with our data-driven test $\phi_{b_n,\delta_n, a_n}^\alpha(\cdot)$ in (\ref{eq_problem_st_7}).  
Here  we focus on the non-asymptotic capacity of our framework to detect the two hypotheses.  
For this, we propose the following detection times:\footnote{To simplify the notation, the dependency of $\mathcal{\tilde{T}}_0((Z_n)_{n\geq 1})$ and $\mathcal{\tilde{T}}_1((Z_n)_{n\geq 1})$ on $(b_m)$, $(\delta_m)$, $(a_m)$  and the scalar parameters $w$, $\alpha$ introduced in Sec.\ref{sub_sec_tsp_algorithm} will be considered implicit.}
\begin{align} \label{eq_sub_I_test_1}
	\mathcal{\tilde{T}}_0((Z_n)_{n\geq 1})  \equiv \sup {\left\{ m\geq 1:  \phi^\alpha_{b_m,\delta_m,a_m}(Z_1,..,Z_m)= 1 \right\}},\\ 
	\label{eq_sub_I_test_1b}
	\mathcal{\tilde{T}}_1((Z_n)_{n\geq 1})  \equiv \sup {\left\{ m\geq 1:  \phi^\alpha_{b_m,\delta_m,a_m}(Z_1,..,Z_m)= 0 \right\}}.
\end{align} 
$\mathcal{\tilde{T}}_0((Z_n)_{n\geq 1})$ and $\mathcal{\tilde{T}}_1((Z_n)_{n\geq 1})$ are random variables (rvs.) in $\mathbb{N}^*=\mathbb{N}\cup  \left\{\infty \right\}$ determining when our 
test reaches $0$ and $1$, respectively.  Indeed, Theorem \ref{th_test_consistency_2}  (under specific conditions) tells us that under $\mathcal{H}_0$,  $\mathbb{P}(\mathcal{\tilde{T}}_0((Z_n)_{n\geq 1}) < \infty)=1$ and that under $\mathcal{H}_1$, $\mathbb{P}(\mathcal{\tilde{T}}_1((Z_n)_{n\geq 1}) < \infty)=1$. However, we are interested in the complete distribution of the rvs. $\mathcal{\tilde{T}}_0((Z_n)_{n\geq 1})$ and $\mathcal{\tilde{T}}_1((Z_n)_{n\geq 1})$ under $\mathcal{H}_0$ and $\mathcal{H}_1$, respectively.  In particular,  we are interested in evaluating the pmf of $\mathcal{\tilde{T}}_i(\cdot)$, i.e., $(\mathbb{P}(\mathcal{\tilde{T}}_i((Z_n)_{n\geq 1}) = k))_{k\geq 1}$ under $H_i$ (with $i\in \left\{0,1 \right\}$). Looking at these distributions, for any  $\epsilon>0$  we can define 
\begin{align}
	\label{eq_sub_I_test_2}
	M_0(\epsilon) &\equiv  \min \left\{ m \geq 1,\  \mathbb{P}(\mathcal{\tilde{T}}_0((Z_n)_{n\geq 1}) \leq m) \geq 1-\epsilon \right\}\\  
	\label{eq_sub_I_test_2b}
	M_1(\epsilon) &\equiv  \min \left\{ m \geq 1,\  \mathbb{P}(\mathcal{\tilde{T}}_1((Z_n)_{n\geq 1}) \leq m) \geq 1-\epsilon \right\}  
\end{align} 
$M_0(\epsilon)$ (and $M_1(\epsilon)$) indicates how many observations (sampling complexity) are needed to detect independence (and statistical dependency) with a confidence probability $1-\epsilon$ under $\mathcal{H}_0$ (and $\mathcal{H}_1$). 
In the context of our solution, 
for a given TSP scheme (function of the parameters $l$, $w$, $\alpha$ and  the sequence $(a_n)$), we will look at the trade-off expressed in the pair $(M_0(\epsilon),M_1(\epsilon))$ of induced tests when varying key parameters of our method. 

\begin{figure*}[ht]
\centering
\includegraphics[width=1.00\textwidth]{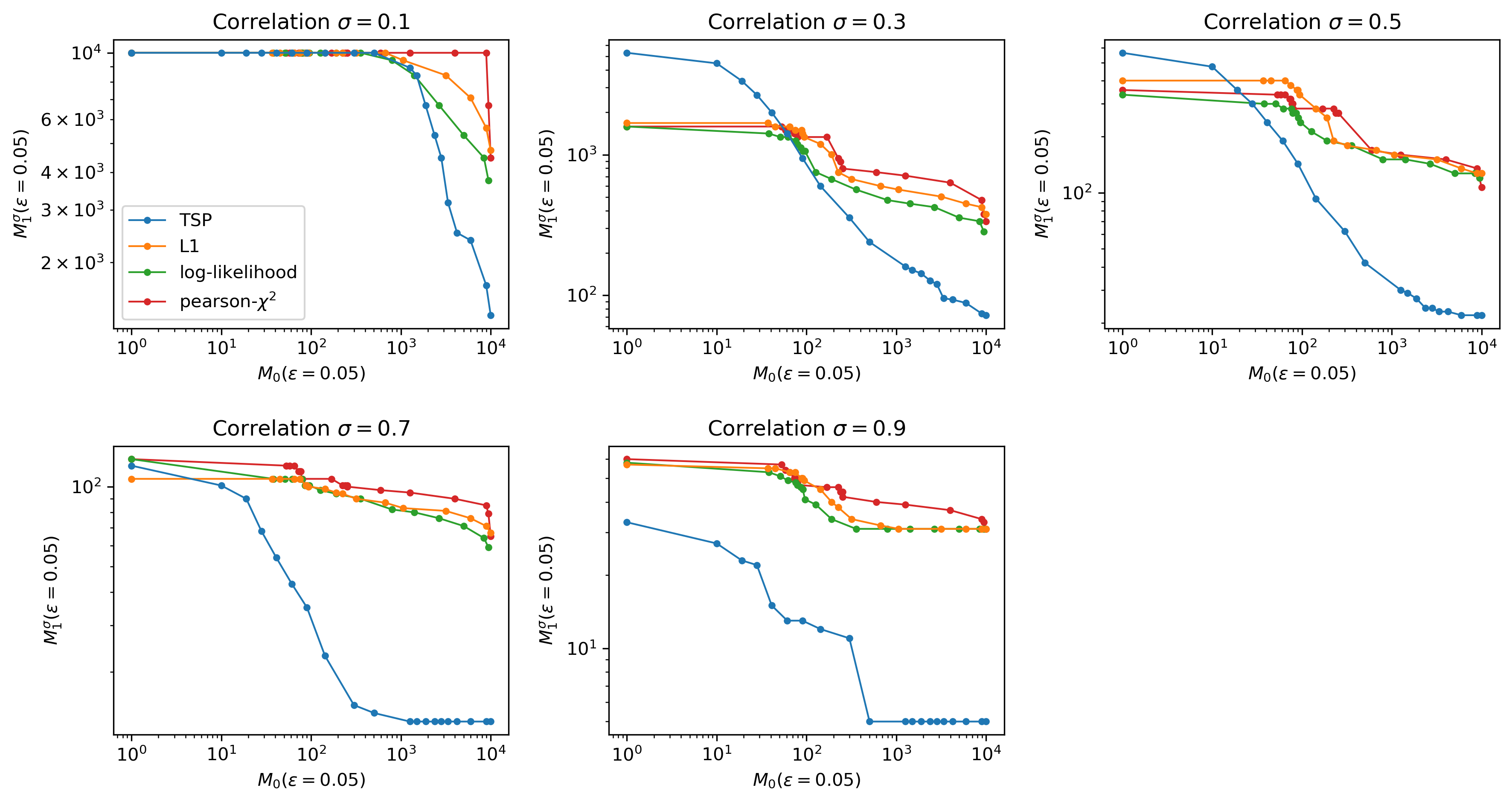}
\caption{Illustration of the trade-off between $M_0(\epsilon)$ and $M^\sigma_1(\epsilon)$ obtained for our TSP test,  the $L_1$-test, the log-likelihood test, and the $\chi^2$-test. The TSP test parameter configuration is $l=0.001$ and $w=0.1$, with $\alpha\in[0.00001,0.0005]$. The $L_1$ test parameter configuration is $m(n)=n^{0.2}$, with $C\in[0.7,1.46]$. The log-likelihood test parameter configuration is $m(n)=n^{0.2}$, with $C\in[0.05,0.3]$. The Pearson-$\chi^2$ test parameter configuration is $m(n)=n^{0.25}$, with $C\in[0.005,0.185]$. Five scenarios of correlation are presented for $\mathcal{H}_1$ considering $\sigma = \mathbb{E}(XY) \in \left\{0.1, 0.3,0.5,0.7, 0.9 \right\}$.}
\label{fig6}
\end{figure*}

\subsubsection{Simulation Setting} 
\label{subsec_simulations_setting}
We consider a joint vector $Z=(X,Y)$ following a zero-mean Gaussian distribution in $\mathbb{R}^2$ where the correlation coefficient determining $I(X,Y)$ is parametrized by $\sigma=\mathbb{E}(XY)$.
Concerning the alternative hypothesis ($\mathcal{H}_1$), we  consider different levels of MI indexed by $\sigma = \mathbb{E}(XY) \in \left\{0.1,0.3,0.5,0.7,0.9 \right\}$.
We use $1,000$ iid realizations of  $(X,Y)$ under the different scenarios ($\mathcal{H}_0$ and $\mathcal{H}^\sigma_1$) to run our test with these iid samples. By doing so, we obtain $1,000$ realizations of the rvs. $\mathcal{\tilde{T}}_0((Z_n)_{n\geq 1})$ and $\mathcal{\tilde{T}}_1((Z_n)_{n\geq 1})$ and with those realizations we obtain an empirical estimation of their pmfs. and empirical estimations of $M_0(\epsilon)$ and $M^\sigma_1(\epsilon)$ (indexed by the value $\sigma$), respectively.\footnote{By the law of large numbers,  the estimators of $M_0(\epsilon)$ and $M^\sigma_1(\epsilon)$ are strongly consistent.  $1,000$ samples shown to be sufficient for the purpose of the analysis.}

\subsubsection{Parameter Selection for $\phi_{b_n,\delta_n, a_n}^\alpha(\cdot)$} 
\label{subsec_parameters_experiments}
To evaluate the sensitivity of our TSP scheme, we consider the following fixed  sequences $(a_n)=(0.5 \cdot n^{-1})$ and $(\delta_n)=(e^{-n^{1/3}})$ in the admissible regime established in Theorem \ref{th_test_consistency_2}.\footnote{$(a_n)_{n\geq 1}$ needs to be $o(1)$ and $(\delta_n)$ needs to be $\ell_1(\mathbb{N})$. Other configurations for $(a_n)$ and $(\delta_n)$ can be explored within the  admissible range declared in Theorem \ref{th_test_consistency_2}.} 
Considering $(b_n)=(w\cdot n^{-l})$ (parametrized by $w$ and $l$) and  $\alpha$ (used to solve $\hat{T}_{b_n,\delta_n} (\alpha)$ in Eq.(\ref{eq_sub_sec_tsp_algorithm_1})),  we consider a preliminary analysis on  MI estimation 
to select a range of reasonable values (in Supplemental Material - Appendix \ref{sub_mi_estimation}). In particular, we consider $l\in \left\{0.001, 0.125, 0.167, 0.25, 0.3 \right\}$ and  $w\in \left\{ 0.005, 0.01, 0.05, 0.1, 0.2 \right\}$. 
We proceed as follows: given fixed parameters $l$ and $w$,
we explore values of  $\alpha$ in  $[0, 4\cdot 10^{-4})$ to express the trade-off between $M_0(\epsilon)$ and $M^\sigma_1(\epsilon)$ under different data scenarios: $\sigma\in \left\{0.3,0.5,0.7 \right\}$.

\subsubsection{Results} 
\label{subsec_results_tests}
Figure \ref{fig4} shows the curves expressing the trade-off between $M_0(\epsilon)$ and $M^\sigma_1(\epsilon)$ for different parameter configurations of our method in the range for $l$ and $w$ mentioned above and for $\epsilon=0.05$. In general, we notice that the effect of these two parameters is relevant in the trade-off expressed in the curves. Both $w$ and $l$ determine the growing phase (i.e., the creation of the full tree) and also the pruning phase (regularization in Eq.(\ref{eq_sub_sec_tsp_algorithm_1})) because $r_{b_n, \delta_n}(\left| T \right|)$ is also a function of $b_n$. Therefore, the effects of $(b_n)_{n\geq 1}$ on the results are not easy to express theoretically.

Describing the curves, we could say in general that a bigger value of $w$ within the explored range reduces the full tree's size in the growing phase. This effect is expressed in a better trade-off of the pair ($M_0(\epsilon)$,  $M^\sigma_1(\epsilon)$) in the regime where $M_0(\epsilon)\leq 10^2$, at the expense of a worse trade-off of ($M_0(\epsilon)$,  $M^\sigma_1(\epsilon)$) when $M_0(\epsilon)\geq 10^3$. A similar general effect in the trade-off ($M_0(\epsilon)$,  $M^\sigma_1(\epsilon)$)  is produced when decreasing the value of $l$ within the explored range. Our family of solutions offers a collection of different trade-offs between $M_0(\epsilon)$ and $M^\sigma_1(\epsilon)$ by exploring different values of $\alpha \in [0, 4\cdot 10^{-4})$ in our solution. Therefore, the selection of the best parameters should be a function of the regime of sample size  we want to consider for the detection of $\mathcal{H}_0$. For the final comparison with  alternative approaches, we decided to consider one of the curves with a less prominent decreasing transition in Figure \ref{fig4} (obtained for $w=0.1$ and $l=0.001$).

\begin{figure*}[ht]
\centering
\includegraphics[width=0.7\textwidth]{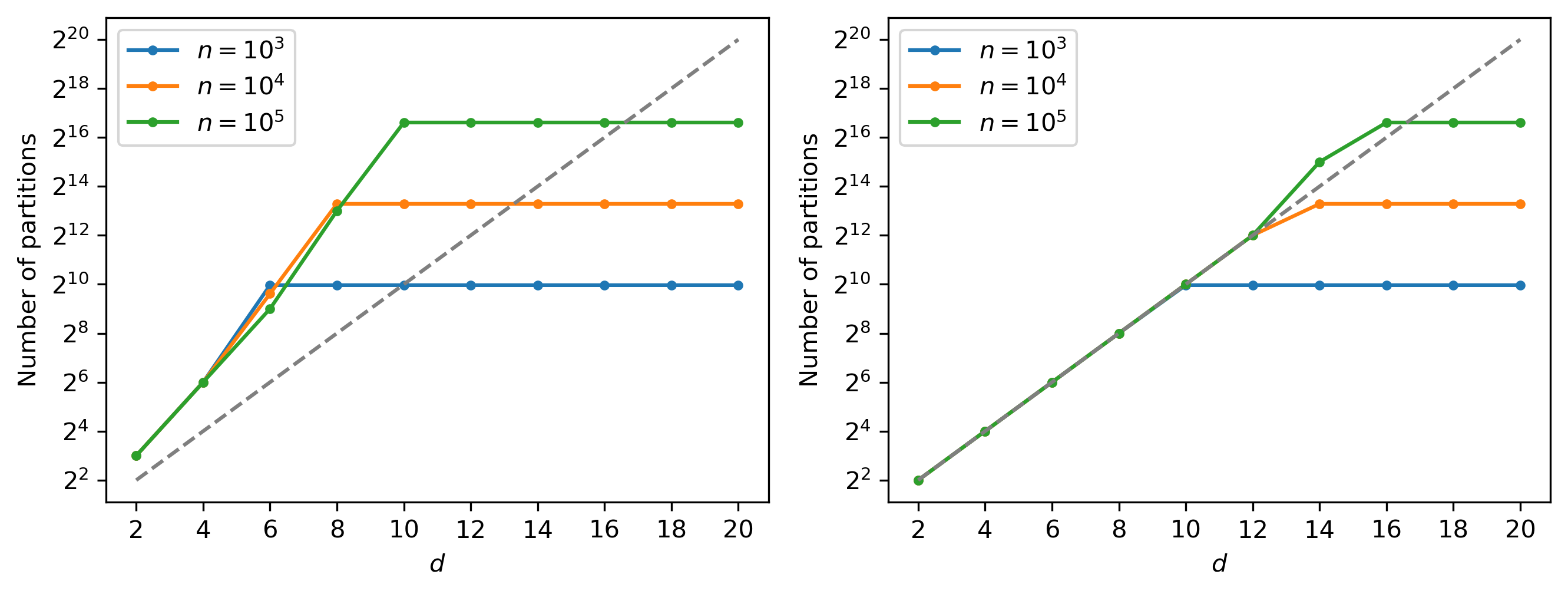}
\caption{
Number of partitions generated for the full tree of the TSP under $l=0.001$ as a function of the dimensionality $d$ of the joint space of the random variables for $n\in\{10^3,10^4,10^5\}$ number of samples and two heuristic rules for the parameter $w$. Left: $w=0.1^{d/2}$. Right: $w=0.225^{d/2}$. Diagonal line determines the minimum number of partitions that allow each dimension of the joint space to be partitioned at least once.}
\label{Fig12}
\end{figure*}

Figure \ref{fig5} illustrates the estimated (empirical) pmfs of $\mathcal{\tilde{T}}_0((Z_n)_{n\geq 1})$ and $\mathcal{\tilde{T}}_1((Z_n)_{n\geq 1})$ 
under $\sigma=0$ ($\mathcal{H}_0$) and $\sigma=0.3$ ($\mathcal{H}_1$), respectively. 
These histograms illustrate how the distributions are affected by the regularization parameter $\alpha$ and, with that, the trade-off in $M_0(\epsilon)$ and $M^\sigma_1(\epsilon)$ observed 
in Figure \ref{fig4}. Specifically, a small value of $\alpha$ concentrates the distribution of $\mathcal{T}_0((Z_n)_{n\geq 1})$ in a high number of observations, while $\mathcal{T}_1((Z_n)_{n\geq 1})$ concentrates on a lower number of observations. This leads to a high value of $M_0(\epsilon)$ and a low value of $M_1^{\sigma}(\epsilon)$ for that specific $\alpha$. The opposite holds for higher values of $\alpha$. 

Finally, we compared our TSP scheme in terms of the trade-off in $(M_0(\epsilon), M^\sigma_1(\epsilon))$ with $\epsilon=0.05$ and $\sigma \in \left\{0.1,0.3,0.5,0.7,0.9 \right\}$ (for $\mathcal{H}_1$) 
exploring three strategies presented in the literature \cite{gretton_2010}: the $L_1$ test, the log-likelihood test, and the Pearson-$\chi^2$ test. Each trade-off curve is generated by multiplying a well-selected range of values $C\in\mathbb{R}$ to the corresponding threshold for the independence detection \cite{gretton_2010}. 
For the parameters of the alternative tests, we consider regimes where these tests are strongly consistent\footnote{We consider the number of partitions
per dimension as $m(n)=n^p$ with $p\in(0,0.5)$, which satisfies the strong-consistency conditions of the $L_1$ and the log-likelihood tests established in \cite{gretton_2010}. Although there is no explicit proof of Pearson-$\chi^2$ test strong-consistency, we use a regime of values for $p \in (0,1/3)$ according to the  range
suggested in \cite{gretton_2010}.}. Within that, we selected parameters that offer a smooth transition between their sampling complexity pairs' trade-off as we did to select the parameters of our method.\footnote{More information is presented in the {Supplemental Material} -  Appendix  \ref{sec_comp}.} 

Figure \ref{fig6} presents these curves for our method and the alternative approaches. These curves express the trade-off between the ability to detect independence and non-product probabilistic structure ($\mathcal{H}_0$  and $\mathcal{H}_1$) from data. In general,   
we could say  that our method shows better results in almost all explored scenarios ($\sigma \in \left\{0.1,0.3,0.5,0.7,0.9 \right\}$) where the TSP trade-off curves dominate the others. There is one exception to this general trend in the regime of $M_0(\epsilon)<10^2$ for lower values of correlation for $\mathcal{H}_1$. 
However, in all the other regimes, our method performs better than the alternatives. In particular, it performs better than its closest relative, which is the log-likelihood test that uses non-adaptive partitions \cite{gretton_2010}. Interestingly, our method's performance improvements increase with the magnitude of $\sigma$ (for $\mathcal{H}_1$). This shows that our approach's advantage is more prominent when the alternative scenario  
has a higher level of mutual information. 

\begin{figure*}[ht]
\centering
\includegraphics[width=0.9\textwidth]{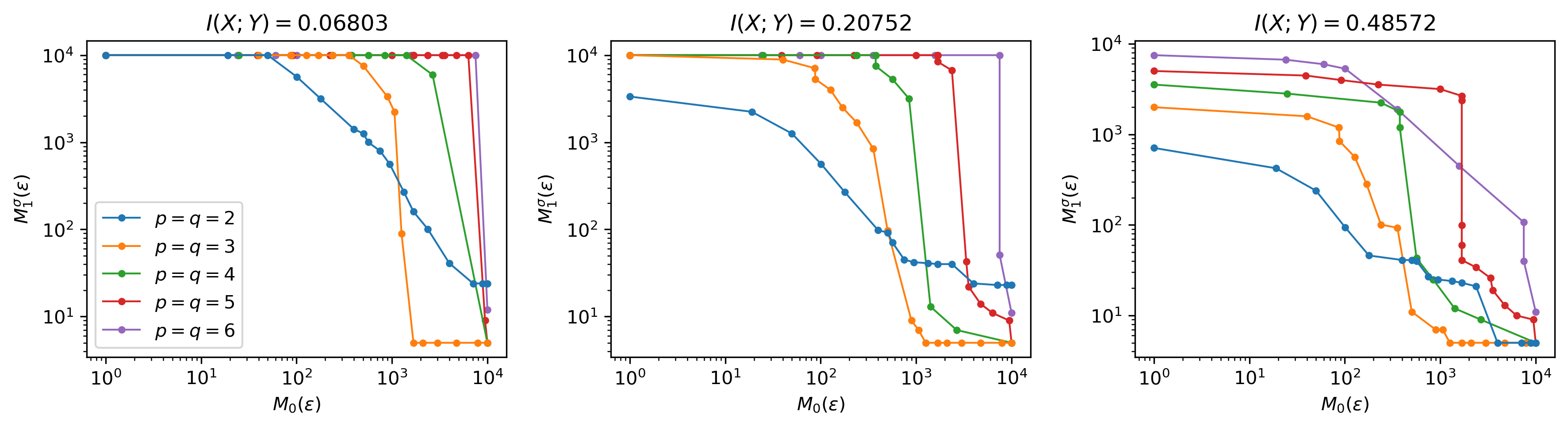}
\caption{
Illustration of the trade-off between $M_0(\epsilon)$ and $M^\sigma_1(\epsilon)$ obtained for the TSP test using samples from a multidimensional Gaussian distribution with component-wise correlation $cov(X_i,Y_j)=\delta_{ij}\sigma(d)$, where $\sigma(d)$ is selected to preserve the mutual information between the variables along the dimensions $p=q\in\{2,3,4,5,6\}$ of each random variable under $l=0.001$ and $w=0.225^{d/2}$ with $d=p+q$. Three scenarios of mutual information are explored $I(X;Y)\in\{0.06803, 0.20752, 0.48572\}$, which correspond to the theoretical mutual information obtained in the univariate Gaussian case $p=q=1$ with $\mathbf{E}(XY)=\sigma, \sigma\in\{0.3,0.5,0.7\}$, respectively.}
\label{Fig13}
\end{figure*}

\subsection{Multidimensional Analysis}
We conclude this analysis by evaluating our method in higher dimensions. As in the previous analysis $d=2$, we need to select the parameters $w$ and $l$ of our scheme. In this multi-dimensional context, we select a value of $l$ that offers a smooth trade-off curve according to the experimental results in Figure \ref{fig4}. However, as $w$ has a stronger impact on the resulting number of partitions, rather than choosing it individually for every dimension $d$, we propose $w=C^{p\cdot d}$ as a heuristic rule for selecting $w$ according to $d$, where $C>0$ and $p>0$. 

This rule and its parameters were designed on the principle that every dimension of the joint space $\mathbb{R}^d$ should be explored at least once in the growing phase. 
This criterion ensures that our data-driven scheme explores every coordinate of $\mathbb{R}^d$ in the pursuit of detecting relevant statistical dependencies under $\mathcal{H}_1$. For a dimension $d>0$, the full tree needs to have at least $2^d$ leaves (cells). For this, a basic condition is that $b_n$ is lower bounded by $1/n$ (our stopping criterion requires at least one sample per cell), which implies that to explore all the $d$ dimensions we need at least $2^d$ i.i.d. samples to meet this requirement. Then assuming the non-trivial regime $n\geq 2^d$, we selected $C$ and $p$ in our rule with the objective that $\left| T^{full}_{b_n} \right| \geq 2^d$ is met independent of the value of $d$ (as long as $n\geq 2^d$). This happens if $w$ becomes smaller when $d$ increases, which happens in our rule for $C<1$ and $p>0$.

Figure \ref{Fig12} shows the number of cells of the full-tree, namely $\left| T^{full}_{b_n} \right|$, for two settings: $w=0.1^{d/2}$ ($C=0.1$ and $p=1/2$) in the left panel, and $w=0.225^{d/2}$ ($C=0.225$ and $p=1/2$) in the right panel. The curves are shown across $d$ for three scenarios of sample-size $n=10^3$, $n=10^4$ and $n=10^5$. We also include the curve $2^d$ (dashed line) to indicate the critical lower bound.
Both settings satisfy the proposed criterion of partitioning each dimension at least once (points above the diagonal) in the whole regime when $n\geq 2^d$. The horizontal line shows the points  where the number of partitions generated by our method is equal to the number of samples (which happens when $b_n=1/n$). Interestingly, when $d$ approaches the critical condition $2^d = n$, our full tree meets a scenario where every cell has one sample. Of the two settings, the selection on the right panel is more conservative, and it is the one we will use for the rest of this analysis. This selection partitions each dimension just above the minimum requirement (for the values of $d$ where $n\geq 2^d$) and achieves the  critical condition $n=2^d$ as $d$ increases.

Using this rule $w=0.225^{d/2}$ and $l=0.001$, we consider a zero-mean joint Gaussian vector $({X},{Y})$ in $\mathbb{R}^{d}$, where $X=(X_1,..,X_p)$ and ${Y}=(Y_1,..,Y_q)$ and, under $\mathcal{H}_1$,  a diagonal co-variance of the form $cov(X_i,Y_j)=\delta_{ij}\sigma(d)^2$ (with $\delta_{ij}$ the Kronecker’s delta).  For these results, we consider $n=10^4$ i.i.d samples meeting the condition that $n\geq 2^d$  for $d\in \left\{4, 6, 8, 10, 12 \right\}$. For the alternative, we consider $\sigma\in\{0.3,0.5,0.7\}$ for $d=2$ ($p=q=1$), and we choose the values of $\sigma(d)$ for $d\geq 4$ such that the mutual information between $X$ and $Y$ is the same as that obtained for $d=2$ with $\sigma\in\{0.3,0.5,0.7\}$. Therefore, for the alternative, we create three scenarios of constant MI $I(X;Y)\in\{0.06803, 0.20752, 0.48572\}$  across dimensions. This experimental design allows us to isolate the effect of dimensionality from the discrimination of the underlying task, which is known to be proportional to the value of MI under $\mathcal{H}_1$ \cite{kullback1958,csiszar_2004,cover_2006}.

Figure \ref{Fig13} presents the trade-off between $(M_0(\epsilon), M^\sigma_1(\epsilon))$ for the following values of 
$\sigma(d)$ (under $\mathcal{H}_1$) and for different dimensions $d\in \left\{4, 6, 8, 10, 12 \right\}$. 
First, we confirm that our scheme detects both hypotheses with a finite sample size as our theoretical results predict. Here,  
we observe that by increasing the dimensions of the problem the curves $(M_0(\epsilon), M^\sigma_1(\epsilon))$ show sharper transitions (beyond a critical value for $M_0(\epsilon)$) and overall the problems become (in general) more difficult: for a given value of $M_0(\epsilon)$, its respective $M^\sigma_1(\epsilon)$ increases with the dimension $d=p+q$. In other words, for a relatively good capacity to detect independence (when $M_0(\epsilon) \leq 10^3$), it is more challenging for our test to detect the alternative as $d$ is higher. This performance trend could be attributed to the observation that in higher dimensions it is more challenging for a non-parametric framework to detect salient features under the alternative $\mathcal{H}_1$ and, consequently, more observations (evidence) are required to correctly detect the alternative  with probability $1-\epsilon$. On the other hand, if we fix a dimension, let us say $d=p+q=8$, the higher the MI the better the trade-off between ($(M_0(\epsilon), M^\sigma_1(\epsilon))$), which is a trend consistent with our observation that MI is an indicator of the task discrimination, and consequently, the performance of our scheme is sensitive to the level of MI under the alternative.

\section{Summary and Final Discussion}
\label{sec_final}
We present a novel framework to address the problem of universal testing of independence using data-driven representations. Consistency and finite length (non-asymptotic) results express our solution's capacity to adapt its representations to the problem's sufficient statistics. This capacity is particularly evident under the hypothesis of independence. Precise results show that our scheme detects this scenario -- collapsing to the trivial partition with one cell for representing the joint space -- with a finite sample size. On the experimental side, our solution offers a computationally efficient implementation. 

In a controlled experimental setting, we show that our scheme offers a clear advantage compared with alternative non-parametric solutions that do not adapt their representations to the data. These results confirm the critical role that data-driven partitions could play in implementing a universal test of independence. 
Further analysis is needed to fully uncover this method's potential for machine learning (ML) problems and other applications. We anticipate that ML algorithms could benefit from both the representations obtained from our solutions (as an approximator of sufficient statistics) and our solution's capacity to detect independence with a finite number of samples.

\subsection{{Limitations and Future Work}}
Our distribution-free results in Section \ref{sec_finite_length} provide a range of admissible parameters for our test; however, they do not provide 
a criterion for selecting them in a specific problem setting.  To address this last practical aspect,  we select a set of parameters from empirical observations and some basic heuristic rules. Then, it is an open research avenue to fully study conditions that would make a selection of optimal parameters (or nearly optimal) for our test under some specific conditions. Along these lines, it is relevant to further investigate specific classes of models,  and based on this model assumption find a more constrained range of good parameters with improved finite-length performance results.
In this vein, there are interesting ideas and results worth exploring about adaptivity that have been explored in statistical learning \cite{bickel_1998} and universal source coding \cite{merhav_1998,bontemps_2014}.

\subsection{{A Recent Related Work}}
The authors in \cite{zhang2021beauty} have recently introduced a non-parametric test 
with a similar oracle-base design principle. They proposed a data-adaptive weight strategy to approximate the ideal weights of a NP statistic. That strategy has an interesting connection with our design approach in Section \ref{sec_regret} because they adapt  key parameters of their method to approximate a best (oracle) {statistic}. On the differences, the authors in \cite{zhang2021beauty} addressed the independence test by performing a binary expansion filtration of the observations and adapting the weights matrix of a quadratic form. Instead, our work addresses the independence test by estimating the mutual information (our statistics) through an adaptive tree-structured partition of the observation space.  Therefore, although both approaches share a similar learning principle, the methods and components used to adapt them to the sample are entirely different.

\subsection{Statistical Significance Analysis}
\label{sec_stat_sig}
We want to conclude with a discussion about the statistical significance of our test. 
Given our test $\phi_{b_n,\delta_n,a_n}(\cdot)$, and assuming  $\mathcal{H}_0$, the statistical significance is expressed by $\alpha(b_n,\delta_n,a_n) \equiv \mathbb{P}(\left\{ \phi_{b_n,\delta_n,a_n}(Z_1,..,Z_n)=1 \right\})$. Importantly, from the proofs of Theorem \ref{th_nearly_minimax_opt_regret}  and Theorem \ref{th_test_consistency_2}, we have the following:
\begin{proposition}\label{pro_bounds_on_alpha_n}
	Under the assumptions of Theorem \ref{th_nearly_minimax_opt_regret} on $(b_n)$ and $(\delta_n)$,
	for any $P \in \mathcal{P}_0$ ($\mathcal{H}_0$) and any $a_n>0$, it follows that
		$\alpha(b_n,\delta_n,a_n) \leq \delta_n$,   $\forall n\geq 1$.\footnote{The proof is presented in the Supplemental Material - Appendix \ref{app_stat_sig}.1).}
\end{proposition}
This result implies that the statistical significance of our test is fully controlled by the sequence $(\delta_n)$, which is one of our design parameters. 
Furthermore, the proof of this result shows that this bound is uniform over the class of models in $\mathcal{P}_0$.
Direct corollaries of Proposition \ref{pro_bounds_on_alpha_n} are the following:
\begin{itemize}
	\item adding the assumptions of Theorem \ref{th_test_consistency_2} and under $\mathcal{H}_0$, it follows that 
	 $\lim_{n \rightarrow  \infty} \alpha(b_n,\delta_n,a_n) =0$. Importantly,  this convergence to zero is uniform for every 
	 model in $\mathcal{P}_0$. [Uniform vanishing condition on $(\alpha(b_n,\delta_n,a_n))_{n \geq 1}$]
	 \item adding the assumptions of Theorem \ref{th_bound_on_tail_of_T0} and under $\mathcal{H}_0$, it follows 
	 that for any $n\geq 1$, $\alpha(b_n,\delta_n,a_n) \leq \delta_n \approx  e^{-n^{1/3}}$. Therefore, we achieve an 
	 exponentially fast vanishing error under $\mathcal{H}_0$. Importantly, this velocity is obtained uniformly (distribution-free) over $P \in \mathcal{P}_0$.
\end{itemize}
These uniform bounds on the statistical significance are obtained under $\mathcal{H}_0$. Under $\mathcal{H}_1$,
we have from strong consistency (Definition \ref{def_consistency}) that $\mathbb{P}(\left\{ \phi_{b_n,\delta_n,a_n}(Z_1,..,Z_n)=1 \right\})$ (the power of the test) tends to $1$ as $n$ tends to infinity.\footnote{This argument is presented in the Supplemental Material - Appendix \ref{app_stat_sig}.2).}

\section{Acknowledgment}
This manuscript is based on work supported by grants of CONICYT-Chile, Fondecyt 1210315, and the Advanced Center for Electrical and Electronic Engineering, Basal Project FB0008.  We thank the anonymous reviewers for providing valuable comments and suggestions that helped to improve the technical quality and organization of this work. Finally, we thank Diane Greenstein for editing and proofreading all this material.

\appendices
\section{Proof of Theorem \ref{th_main}}
\label{proof_th_main}
\begin{proof} 
	In general, first we know  that if $(\phi_n(Z^n_1))_{n\geq 1}$ reaches $0$ eventually with probability one, it is equivalent to saying that the process $(\phi_n(Z^n_1))_{n\geq 1}$ does not visit the  event $ \left\{ 1 \right\}$ infinitely often (i.o.). This observation reduces to verify the following
	\begin{align}\label{eq_problem_st_1}
		&\mathbb{P}\left(\lim\sup_n \left\{z^n_1\in \mathbb{R}^{d n}: \phi_n(z^n_1)=1 \right\} \right) \equiv \nonumber\\ 
		&\mathbb{P}\left(\bigcap_{m \geq 1} \bigcup_{n \geq m} \left\{z^n_1\in \mathbb{R}^{d n}: \phi_n(z^n_1)=1 \right\} \right) =   0.
	\end{align}
	Equivalently, saying that $(\phi_n(Z^n_1))_{n\geq 1}$ reaches $1$ eventually with probability one reduces to 
	\begin{align}\label{eq_problem_st_2}
		&\mathbb{P}\left(\lim\sup_n \left\{z^n_1\in \mathbb{R}^{d n}: \phi_n(z^n_1)=0 \right\} \right) \equiv\nonumber\\
		&\mathbb{P}\left(\bigcap_{m \geq 1} \bigcup_{n \geq m} \left\{z^n_1\in \mathbb{R}^{d n}: \phi_n(z^n_1)=0 \right\} \right) =   0.
	\end{align}

	For the rest of the proof, we use $O^n_1,..,O^n_n$ to denote $O_{\pi_n}(Z_1),....,O_{\pi_n}(Z_n)$ and $o^n_1,..,o^n_n$ 
	to denote $O_{\pi_n}(z_1),....,O_{\pi_n}(z_n)$. 

	Under $\mathcal{H}_0$, we have from the third hypothesis that  ${\hat{i}_{\pi_{n}}(O^n_1,....,O^n_n)}/{a_n}$
	tends to $0$ with probability one, which means that for any $\epsilon>0$
	\begin{equation}\label{eq_proof_1_6}
			\mathbb{P}\left(\bigcap_{m \geq 1} \bigcup_{n \geq m} B^{\pi_n, a_n}_{\epsilon,n}\right) = 0,
	\end{equation}
	where $B_{\epsilon,n}^{\pi_n, a_n} \equiv \left\{z^n_1\in \mathbb{R}^{nd}: \hat{i}_{\pi_{n}}(o^n_1,....,o^n_n) > \epsilon \cdot a_n  \right\}$.
	Under $\mathcal{H}_0$,  let us consider the error event of $\phi_{\pi_n, a_n}()$ by
	$$E_n^{\pi_n, a_n} \equiv \left\{ z^n_1 \in \mathbb{R}^{nd}: \phi_{\pi_n, a_n}(o^n_1,..,o^n_n) =1 \right\}.$$   
	If $z^n\in E_n^{\pi_n, a_n} $ (by definition of the rule $\phi_{\pi_n, a_n}(\cdot)$), it follows that $\hat{i}_{{\pi_n}}(o^n_1,....,o^n_n) \geq a_n$ (see (\ref{eq_problem_st_7})). Then from definition of $B_{\epsilon,n}^{\pi_n, a_n}$,  for any $\epsilon<1$ it follows that $E_n^{\pi_n, a_n} \subset B_{\epsilon,n}^{\pi_n, a_n}$. This implies that 
	$\bigcap_{m \geq 1} \bigcup_{n \geq m} E_n^{\pi_n, a_n} \subset \bigcap_{m \geq 1} \bigcup_{n \geq m} B_{\epsilon,n}^{\pi_n, a_n}$ which from (\ref{eq_proof_1_6}) implies that 
	\begin{equation}\label{eq_proof_1_7}
			\mathbb{P}\left(\bigcap_{m \geq 1} \bigcup_{n \geq m}E_n^{\pi_n, a_n}\right) = 0.
	\end{equation}
	
	On the other hand, under the assumption that $I(X,Y)>0$ ($\mathcal{H}_1$), let us choose $\epsilon_0\in (0,I(X,Y))$. Then, from the second assumption (see Definition \ref{def_consistency_suff_stat}), 
	it follows that $\hat{i}_{\pi_{n}}(O^n_1,....,O^n_n)$ convergences to something strictly greater than 
	$\epsilon_0$ with probability one. This formally means that
	\begin{equation}\label{eq_proof_1_8}
		\mathbb{P} \left(   \bigcap_{m \geq 1} \bigcup_{n \geq m} \left\{ z^n_1 \in  \mathbb{R}^{nd}: \hat{i}_{\pi_{n}}(o^n_1,....,o^n_n) \in (0,\epsilon_0) \right\} \right)  = 0,
	\end{equation}
	The error event  of $\phi_{\pi_n, a_n}()$ under $\mathcal{H}_1$ in this case is
	\begin{equation}\label{eq_proof_1_9}
	\tilde{E}_n^{\pi_n, a_n} \equiv \left\{ z^n_1 \in \mathbb{R}^{nd}: \hat{i}_{\pi_{n}}(o^n_1,....,o^n_n) <a_n \right\}.
	\end{equation}
	Finally using the condition that $(a_n)_{n\geq 1}$ tends to zero with $n$ ({$(a_n)_{n}$ is $o(1)$}), we have that 
	\begin{equation}\label{eq_proof_1_10}
	\tilde{E}_n^{\pi_n, a_n} \subset  \left\{ z^n_1 \in \mathbb{R}^{nd}: \hat{i}_{\pi_{n}}(o^n_1,....,o^n_n) \in (0,\epsilon_0) \right\} 
	\end{equation}
	eventually in $n$,  which implies from (\ref{eq_proof_1_8}) that
	\begin{equation}\label{eq_proof_1_11}
			\mathbb{P}\left(\bigcap_{m \geq 1} \bigcup_{n \geq m}\tilde{E}_n^{\pi_n, a_n}\right) = 0.
	\end{equation}
\end{proof}	

\section{Proof of Lemma \ref{th_consistency_regret}}
\label{proof_th_consistency_regret}
We use the following results from \cite{silva_2012}.
\begin{lemma} \cite[Th.3]{silva_2012}
	\label{lemma_consistency_mi}
	Under the assumptions of Lemma \ref{th_consistency_regret}, 
		$\lim_{n \rightarrow \infty} \hat{i}_{\pi_{b_n,\delta_n}}(O_1,....,O_n)= I(X,Y)$ 
	with probability one. 
\end{lemma}
\begin{lemma} \cite[Th.4]{silva_2012}
	\label{lemma_rate_under_independence}
	Under the assumptions of Lemma \ref{th_consistency_regret} part ii),  for any $p>0$,
	$\hat{i}_{\pi_{b_n,\delta_n}}(O_1,....,O_n)$ is $o(n^{-p})$.   \footnote{
	This means formally that $\lim_{n \rightarrow \infty} \frac{\hat{i}_{\pi_{b_n,\delta_n}}(O_1,....,O_n)}{n^{-p}}=0$
	with probability one.} 
\end{lemma}

\begin{proof} Eq.(\ref{eq_main_results_2}) follows from  Lemma \ref{lemma_consistency_mi},  
the fact that by the strong law of large number $\lim_{n \rightarrow \infty} i_n(Z_1,...,Z_n)= I(X,Y)$ with probability one and 
the union bound. 

Under the assumption that $P\in \mathcal{P}_0$ (i.e. $I(X,Y)=0$), we have that $P=Q^*(P)$. This implies by definition that $i_n(Z_1,...,Z_n)=0$ with probability one. Then the regret is basically our empirical information term $\hat{i}_{\pi_{b_n,\delta_n}}(O_1,....,O_n)$. Then, Lemma \ref{lemma_rate_under_independence} implies the result in (\ref{eq_main_results_3}).
\end{proof}

\section{Proof of Theorem \ref{th_nearly_minimax_opt_regret}}
\label{proof_th_nearly_minimax_opt_regret}
\begin{proof}
If we define the event $\mathcal{E}^{n,k}_{\delta_n, b_n} \equiv$
\begin{equation}\label{eq_th_nearly_minimax_opt_regret_2}
  \left\{ z^n_1: 
  \sup_{T \in \mathcal{G}^k_{b_n}} \left|{ D_{\sigma({\pi_T})}({P}||{Q}^*)  - D_{\sigma({\pi_T})}(\hat{P}_n||\hat{Q}_n^*) }\right| \leq  r_{b_n, \delta_n}(k) \right\}, 
\end{equation}
for all $k\in  \left\{1,..,, \left| T^{full}_{b_n} \right| \right\}$, we have that the conditions stated on 
$(b_n)$ and $(\delta_n)$ are the weakest to obtain from (\ref{eq_sub_sec_reim_2}) that $\lim_{n \rightarrow \infty} \sup_{k\in \left\{1,..,, \left| T^{full}_{b_n} \right| \right\}} r_{b_n, \delta_n}(k)=0$. This last condition is crucial to being able to apply Lemma \ref{th_est_err_bound} in $ \left\{\mathcal{E}^{n,k}_{\delta_n, b_n},k \right\}$. In fact, $\mathbb{P} \left( \mathcal{E}^n_{\delta_n, b_n} \equiv \bigcap_{k=1}^{\left| T^{full}_{b_n} \right|} \mathcal{E}^{n,k}_{\delta_n,b_n} \right) \geq 1- \delta_n$ by definition of $r_{b_n, \delta_n}(k)$ in (\ref{eq_sub_sec_reim_3}) and the condition expressed in (\ref{eq_sub_sec_reim_2}).

Importantly under $\mathcal{H}_0$ (i.e., $I(X; Y)=0$), for $k=1$, we can consider that $r_{b_n, \delta_n}(k)=0$,  because $D_{\sigma({\left\{ \mathbb{R}^d \right\} })}({P}||{Q}^*) = D_{\sigma({  \left\{  \mathbb{R}^d \right\} })}(\hat{P}_n||\hat{Q}_n^*)=0$ for any $z^n_1\in \mathbb{R}^{dn}$.   
Consequently,  for any 
$z^n_1\in \mathcal{E}^n_{\delta_n, b_n} = \bigcap_{k=1}^{\left| T^{full}_{b_n} \right|} \mathcal{E}^{n,k}_{\delta_n,b_n}$,  it follows that 
\begin{align}\label{eq_th_nearly_minimax_opt_regret_3}
	 &D_{\sigma({\pi_{\hat{T}_{b_n,\delta_n}}})}(\hat{P}_n||\hat{Q}_n^*)  - r_{b_n, \delta_n}\left(\left| \hat{T}_{b_n,\delta_n} \right|\right) 
	  \geq \nonumber\\
	  &\sup_{T \in \mathcal{G}^k_{b_n}} D_{\sigma({\pi_{{T}}})}(\hat{P}_n||\hat{Q}_n^*)  - r_{b_n, \delta_n}(k )\\
	 \label{eq_th_nearly_minimax_opt_regret_3b}
	 &\geq \sup_{T \in \mathcal{G}^k_{b_n}} D_{\sigma({\pi_{{T}}})}({P}|| {Q}^*)  - 2 r_{b_n, \delta_n}( k), 
\end{align}
for any $k\in  \left\{1,..,  \left| T^{full}_{b_n} \right| \right\}$. The first inequality in (\ref{eq_th_nearly_minimax_opt_regret_3}) is from the definition of $\hat{T}_{b_n,\delta_n}$ in (\ref{eq_sub_sec_reim_4})  and the second inequality in (\ref{eq_th_nearly_minimax_opt_regret_3b}) is from the fact that 
if $z^n_1\in \mathcal{E}^{n,k}_{\delta_n,b_n}$ (see  (\ref{eq_th_nearly_minimax_opt_regret_2})), then
\begin{align}\label{eq_th_nearly_minimax_opt_regret_4}
	 \left| \sup_{T \in \mathcal{G}^k_{b_n}} D_{\sigma({\pi_T})}({P}||{Q}^*)  -  \sup_{T \in \mathcal{G}^k_{b_n}} D_{\sigma({\pi_T})}(\hat{P}_n||\hat{Q}_n^*) \right|  \leq  r_{b_n, \delta_n}(k).
\end{align}
On the other hand, if $z^n_1\in \mathcal{E}^n_{\delta_n, b_n}$, then 
\begin{align}\label{eq_th_nearly_minimax_opt_regret_5}
&D_{\sigma({\pi_{\hat{T}_{b_n,\delta_n}}})}(\hat{P}_n||\hat{Q}_n^*)  - r_{b_n, \delta_n}(\left| \hat{T}_{b_n,\delta_n} \right|) \nonumber\\
&\leq D_{\sigma({\pi_{\hat{T}_{b_n,\delta_n}}})}({P}|| {Q}^*)  \leq D ({P}|| {Q}^*).
\end{align}
At this point, we use the independence assumption\footnote{Under $\mathcal{H}_0$, we have that $D_{\sigma({\pi_T})}({P}||{Q}^*) =0$ for any $T\ll T^{full}_{b_n}$. } and  the two inequalities (\ref{eq_th_nearly_minimax_opt_regret_3b}) and (\ref{eq_th_nearly_minimax_opt_regret_5}) to obtain that  for any $z^n_1\in \mathcal{E}^n_{\delta_n, b_n}$:  
\begin{align}\label{eq_th_nearly_minimax_opt_regret_6}
D_{\sigma({\pi_{\hat{T}_{b_n,\delta_n}}})}(\hat{P}_n||\hat{Q}_n^*)  - r_{b_n, \delta_n}(\left| \hat{T}_{b_n,\delta_n} \right|)=0. 
\end{align}
Finally using that $r_{b_n, \delta_n}(1)=0$, it is simple to verify that for any $z^n_1\in \mathcal{E}^n_{\delta_n, b_n}$ the trivial tree (with one cell) is a solution for  (\ref{eq_sub_sec_reim_4}), i.e., $  \left|  \hat{T}_{b_n,\delta_n} \right| =1$. Then from (\ref{eq_th_nearly_minimax_opt_regret_6}), $D_{\sigma({\pi_{\hat{T}_{b_n,\delta_n}}})}(\hat{P}_n||\hat{Q}_n^*)=\hat{i}_{\pi_{b_n,\delta_n}}(O_{\pi_{b_n,\delta_n}}(Z_1),....,O_{\pi_{b_n,\delta_n}}(Z_n))=0$,
which concludes the argument using that  $\mathbb{P} \left( \mathcal{E}^n_{\delta_n, b_n} \right)  \geq 1- \delta_n$ and the assumption that $(\delta_n)$ is $o(1)$.

Under  $\mathcal{H}_1$ (i.e., $I(X; Y)>0$), Silva {\em et al.} \cite[Th. 2, Eq.(33)]{silva_2012} showed 
that for any $z^n_1\in \mathcal{E}^n_{\delta_n, b_n}$ 
\begin{align*} 
&I(X,Y) -									\hat{i}_{\pi_{b_n,\delta_n}}(O_{\pi_{b_n,\delta_n}}(Z_1),....,O_{\pi_{b_n,\delta_n}}(Z_n)) \nonumber\\
& \leq \min_{T\ll T^{full}_{b_n}} \left[ I(X;Y) -  D_{\sigma({\pi_T})}({P}||{Q}^*(P))  \right] + 2 r_{b_n, \delta_n}(T).
\end{align*}
Again using that $\mathbb{P} \left( \mathcal{E}^n_{\delta_n, b_n} \right)  \geq 1- \delta_n$, 
we obtain the bound in (\ref{eq_main_results_1}).
 \end{proof}

\section{Proof of Lemma \ref{th_structural_detection_ind}}
\label{proof_th_structural_detection_ind}
\begin{proof}  Let us use the definition of the typical set introduced in (\ref{eq_th_nearly_minimax_opt_regret_2}) and 
$\mathcal{E}^n_{\delta_n, b_n} \equiv \bigcap_{k=1}^{\left| T^{full}_{b_n} \right|} \mathcal{E}^{n,k}_{\delta_n,b_n}$. Under the assumption of this result, we know that
 $\mathbb{P} \left( \mathcal{E}^n_{\delta_n, b_n} \right) \geq 1- \delta_n$. 
In addition, in the proof of Theorem \ref{th_nearly_minimax_opt_regret}, it is shown that if $z^n \in \mathcal{E}^n_{\delta_n, b_n}$,
then $\left| \hat{T}_{b_n,\delta_n} \right| =1$, which means that $\pi_{b_n\delta_n}=\left\{\mathbb{R}^d \right\}$.  Therefore, we have that\footnote{Here we use the notation $\pi_{b_n, \delta_n}(z^n)$ to make explicit that the partition is data-driven.} 
\begin{align}\label{eq_th_structural_detection_ind_1}
\mathcal{E}^n_{\delta_n, b_n} \subset \left\{ z^n\in \mathbb{R}^{dn}: \pi_{b_n, \delta_n}(z^n) = \left\{\mathbb{R}^d \right\}  \right\}.
\end{align}
Using the definition of $\mathcal{T}_0((z_n)_{n\geq 1})$, we have that 
\begin{align} \label{eq_th_structural_detection_ind_2}
	&\mathbb{P}(\mathcal{T}_0((Z_n)_{n\geq 1})\leq M) = \mathbb{P}_{{\bf Z}} \left( \bigcap_{n>M}  \left\{z^n: \pi_{b_n, \delta_n}(z^n) = \left\{\mathbb{R}^d \right\} \right\} \right)\\	
		\label{eq_th_structural_detection_ind_2b}
										     &= 1- \mathbb{P}_{{\bf Z}} \left( \bigcup_{n>M}  \left\{z^n: \pi_{b_n, \delta_n}(z^n) \neq \left\{\mathbb{R}^d \right\} \right\} \right)\\
		\label{eq_th_structural_detection_ind_2c}
		&\geq 1- \sum_{n>M}  \mathbb{P}_{{\bf Z}} \left( \left\{z^n: \pi_{b_n, \delta_n}(z^n) \neq \left\{\mathbb{R}^d \right\} \right\} \right)\\
		\label{eq_th_structural_detection_ind_2d}
		&\geq 1- \sum_{n>M}  \mathbb{P}_{{\bf Z}} \left( (\mathcal{E}^n_{\delta_n, b_n})^c \right)  \geq 1- \sum_{n>M} \delta_n. 
\end{align}
The equality in (\ref{eq_th_structural_detection_ind_2}) comes from definition in (\ref{eq_sub_sec_structural_detection_of_H0_1}). The inequality in (\ref{eq_th_structural_detection_ind_2c}) comes from the sub-additivity of $\mathbb{P}_{{\bf Z}}$ and the bounds in (\ref{eq_th_structural_detection_ind_2d}) follow from (\ref{eq_th_structural_detection_ind_1}) and the construction of $\mathcal{E}^n_{\delta_n, b_n}$. Finally, using the assumption that $(\delta_n)\in \ell_1(\mathbb{N})$ in (\ref{eq_th_structural_detection_ind_2d}), we have that 
\begin{align} \label{eq_th_structural_detection_ind_3}
	\mathbb{P}(\mathcal{T}_0((Z_n)_{n\geq 1})<\infty) = \lim_{M \rightarrow \infty}  \mathbb{P}(\mathcal{T}_0((Z_n)_{n\geq 1})\leq M) = 1. 
\end{align} 
\end{proof}

\section{Proof of Theorem \ref{th_test_consistency_2}}
\label{proof_th_test_consistency_2}
\begin{proof}
	Under $\mathcal{H}_1$,  the assumptions on $(b_n)$ and $(\delta_n)$ are within the admissible range stated in Theorem \ref{th_consistency_regret} part i).  Consequently,  we have consistency on the regret,  i.e.,  
	$\lim_{n \rightarrow \infty}\left| i_n(Z_1,...,Z_n) - \hat{i}_{\pi_{b_n,\delta_n}}(O_{\pi_{b_n,\delta_n}}(Z_1),....,O_{\pi_{b_n,\delta_n}}(Z_n) \right| =0$, 
	$\mathbb{P}$-almost surely.  This result is sufficient to obtain that $(\phi_{{b_n,\delta_n}, a_n} (Z_1,....,Z_n))_{n\geq 1}$ reaches $1$ eventually with probability one, from the arguments presented in the proof of Theorem \ref{th_main}.
	
	Under $\mathcal{H}_0$, it is useful to define the following last exit time associated with the detection of independence: 
	\begin{align} \label{eq_proof_th_test_consistency_1}
		&\mathcal{T}((z_n)_{n\geq 1}, (a_n)_{n\geq 1}) \equiv \nonumber\\
		&\sup {\left\{ m\geq 1:  \phi_{b_m,\delta_m,a_m}(z_1,..,z_m)= 1 \right\}} \in \mathbb{N}\cup  \left\{\infty \right\}.
	\end{align}	
	Saying that $(\phi_{{b_n,\delta_n}, a_n} (Z_1,....,Z_n))_{n\geq 1}$ reaches $0$ eventually with probability one is equivalent to 
	the condition that 
	\begin{equation} \label{eq_proof_th_test_consistency_2}
	\mathbb{P}(\mathcal{T}((Z_n)_{n\geq 1}, (a_n)_{n\geq 1})<\infty)=1,  
	\end{equation}	
	from the arguments presented in the proof of Theorem \ref{th_main}. At this point, it is important to revisit the definition of 
	$\mathcal{T}_0((z_n)_{n\geq 1})$ in (\ref{eq_sub_sec_structural_detection_of_H0_1}), where if for some $z^m\in \mathbb{R}^{dm}$ we have that $\left| \hat{T}_{b_m,\delta_m} \right| = 1$; this implies that $\hat{i}_{\pi_{b_m,\delta_m}}(O_{\pi_{b_m,\delta_m}}(z_1),....,O_{\pi_{b_m,\delta_m}}(z_m)= i_m(z_1,...,z_m)=0$ (details on Appendix \ref{proof_th_nearly_minimax_opt_regret}), and therefore, $\phi_{b_m,\delta_m,a_m}(z_1,..,z_m)= 0$ independent of $a_m$ (see Eq.(\ref{eq_problem_st_7})).  Consequently,  we have that
	\begin{align} \label{eq_proof_th_test_consistency_3}
			 &\left\{z^m \in \mathbb{R}^{dm}: \pi_{b_m, \delta_m}(z^m) = \left\{\mathbb{R}^d \right\} \right\}  = \left\{z^m: \left| \hat{T}_{b_m,\delta_m} \right| = 1 \right\}   \nonumber\\
			 &  \subset \left\{z^m \in \mathbb{R}^{dm}:  \phi_{b_m,\delta_m,a_m}(z_1,..,z_m))= 0 \right\}
	\end{align}	
	for any $a_n$,  and it follows that
	\begin{align} \label{eq_proof_th_test_consistency_4}
		&\mathbb{P}(\mathcal{T}((Z_n)_{n\geq 1}, (a_n)_{n\geq 1}) \leq M) =\nonumber\\ 
		&\mathbb{P}_{{\bf Z}} \left( \bigcap_{n>M}  \left\{z^n:  \phi_{b_n,\delta_n,a_n}(z_1,..,z_n)) = 0 \right\} \right)\\
						\label{eq_proof_th_test_consistency_4b}
						&\geq \mathbb{P}_{{\bf Z}} \left( \bigcap_{n>M}  \left\{z^n: \pi_{b_n, \delta_n}(z^n) = \left\{\mathbb{R}^d \right\} \right\} \right)\\
						\label{eq_proof_th_test_consistency_4c}
						&= \mathbb{P}(\mathcal{T}_0((Z_n)_{n\geq 1})\leq M). 	
	\end{align}
	The identity in (\ref{eq_proof_th_test_consistency_4}) follows from the definition in (\ref{eq_proof_th_test_consistency_1}), and equations (\ref{eq_proof_th_test_consistency_4b}) and  (\ref{eq_proof_th_test_consistency_4c}) follow from (\ref{eq_proof_th_test_consistency_3}) and (\ref{eq_th_structural_detection_ind_2}), respectively.  The argument concludes from Lemma \ref{th_structural_detection_ind},  as we know that 
	under the assumptions stated on $(b_n)$ and $(\delta_n)$, $\lim_{M \rightarrow \infty}  \mathbb{P}(\mathcal{T}_0((Z_n)_{n\geq 1})\leq M) = 1$, which implies the result in (\ref{eq_proof_th_test_consistency_2}) from (\ref{eq_proof_th_test_consistency_4c}). 
\end{proof}

\section{Proof of Theorem \ref{th_bound_on_tail_of_T0}}
\label{proof_th_bound_on_tail_of_T0}
\begin{proof}
	Under $\mathcal{H}_0$, we can adopt directly the bound presented in (\ref{eq_th_structural_detection_ind_2d}): 
	\begin{equation} \label{eq_proof_th_bound_on_tail_of_T0_1}
		\mathbb{P}(\mathcal{T}_0((Z_n)_{n\geq 1}) < m) \geq 1- \sum_{n\geq m} \delta_n= 1- \sum_{n\geq m} C e^{-n^{1/3}},
	\end{equation}
	where we include the assumption that $(1/\delta_n)_n \approx (e^{n^{1/3}})$ (i.e., $C$ is positive constant). For the series in (\ref{eq_proof_th_bound_on_tail_of_T0_1}) we have 
	\begin{align} \label{eq_proof_th_bound_on_tail_of_T0_2}
			\sum_{n \geq m} e^{-n^{1/3}}  &= e^{-m^{1/3}}  \left[1+ \frac{e^{-(m+1)^{1/3}} }{ e^{-m^{1/3}} }+...+ \frac{ e^{-(m+j)^{1/3}}} { e^{-m^{1/3}} }+.. \right]\\
							      &= e^{-m^{1/3}} \cdot  \underbrace{\sum_{j\geq 0} e^{-j^{1/3}} }_{\equiv \mathcal{I}}, 
	\end{align}
	where 
	$\mathcal{I}<\infty$. Therefore,  we have that $\mathbb{P}(\mathcal{T}_0((Z_n)_{n\geq 1}) \geq m) \leq C  \mathcal{I} e^{-m^{1/3}}$. 
\end{proof}

\bibliographystyle{IEEEtran}				
\bibliography{main_jorge_silva}				

\newpage
\section{Suplemental Material}
In Section \ref{sub_mi_estimation}, the estimation of mutual information (MI) based on our tree-structured partition (TSP) approach is evaluated. Section \ref{sec_comp} details the selection of parameters of the $L_1$ test, the log-likelihood test, and the Pearson-$\chi^2$ test. These results were used to select the final curves presented in {\bf Figure 3}.  Section \ref{sec_add_exp} complements the empirical results of our method on a heavy-tailed distribution and a multi-modal distribution. Finally, Section \ref{app_stat_sig} presents some results regarding the statistical significance of our test.

\subsection{Preliminary Analysis of Mutual Information Estimation}
\label{sub_mi_estimation}
Analyzing our tree-structured partition (TSP) scheme's capacity to estimate MI in the non-asymptotic regime is important because it provides an indication of the trade-off between the two types of errors that we will encounter when analyzing the performance of its induced test. Here, we present a basic analysis on the effects of the parameters of our scheme:  $l$ and $w$ ($b_n= w\cdot n^{-l}$),  and $\alpha>0$  within the admissible regime provided by our results in {\bf Section VIII} of the main content for $(b_n)_{n\geq 1}$ and $(\delta_n)_{n\geq 1}$. For this analysis, we consider a joint vector $Z=(X,Y)$ following a zero-mean Gaussian distribution in $\mathbb{R}^2$  where the correlation coefficient determining $I(X,Y)$ is parametrized by $\sigma =\mathbb{E}(XY)$.  We evaluate empirically how the estimator $\hat{i}_{b_n,\delta_n}^\alpha(\cdot)$ achieves the true value as $n$ grows and the effects of the mentioned parameters under $\mathcal{H}_0$ ($\sigma=0$) and $\mathcal{H}_1$ ($\sigma>0$). For illustration purposes,  we will consider $\sigma=0.7$ as a relatively high MI regime under $\mathcal{H}_1$.\footnote{Similar analyses where conducted under different level of correlations (under $H_1$) not reported here for brevity.} 

\begin{figure}
\centering
\includegraphics[width=1.00\textwidth]{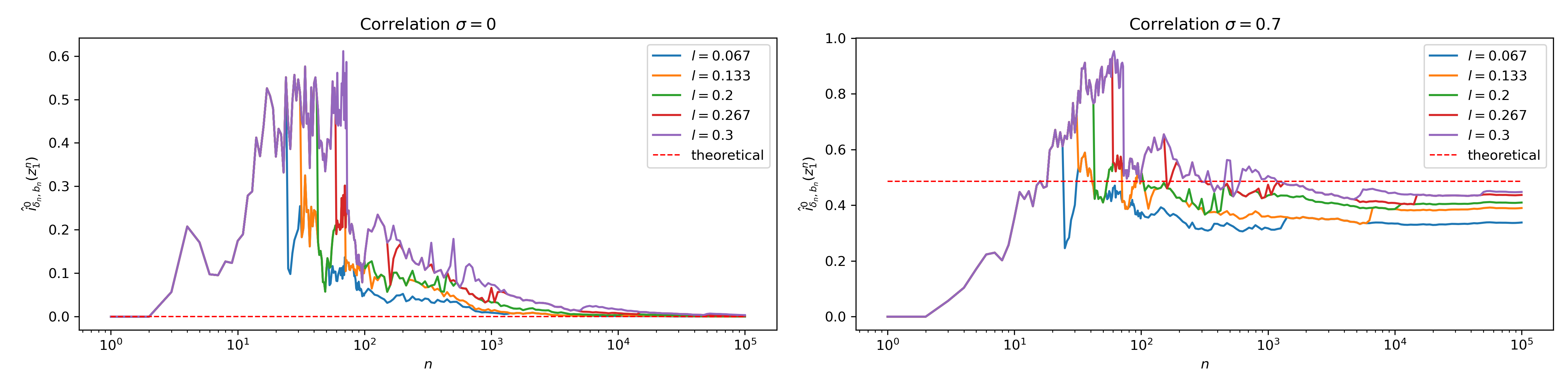}
\caption{Examples of a realization of $\hat{i}_{b_n,\delta_n}^{\alpha=0}(Z^n_1)$  function of the sampling size $n$ under different values of the parameter $l\in  \left\{0.067,0.133, 0.2, 0.267,0.3 \right\}$ with $w=0.05$ and $\alpha=0$ ($b_n=w\cdot n^{-l}$ ). For these results, two scenarios of correlation between $(X,Y)$ are considered: $\sigma=0$ and $\sigma=0.7$. }
\label{fig1}
\end{figure}

To begin, Figure \ref{fig1} shows effects of $l\in (0,1/3)$ fixing $w=0.05$ and $\alpha=0$. Under $\alpha=0$, we look at the performance of the full tree $T^{full}_{b_n}$, which provides the baseline capacity of our regularized solution in {\bf Eq.(17)} of the main content.  \mbox{Figure \ref{fig1}} shows that as $l$ increases (approaching its less conservative value $1/3$ associated with the biggest full tree in the growing phase), the capacity of our scheme to estimate the true MI under $\mathcal{H}_1$ is superior, as expected. On the contrary, the estimation of the true MI under $\mathcal{H}_0$ is better for smaller values of $l$, although beyond $1.000$ samples, the specific value of $l$ is not that critical under $\mathcal{H}_0$. 

\begin{figure}
\centering
\includegraphics[width=1.00\textwidth]{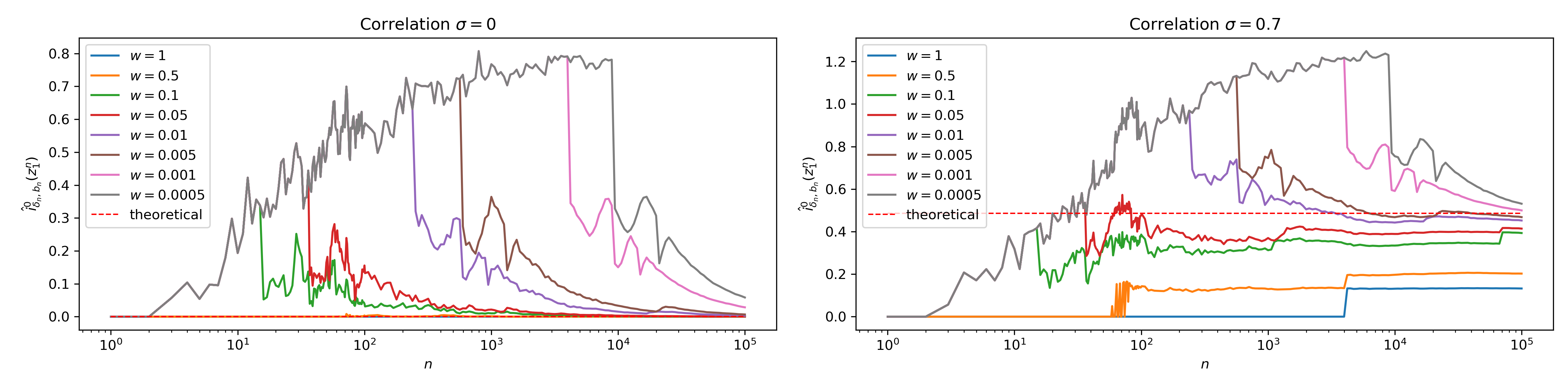}
\caption{Examples of a realization of $\hat{i}_{b_n,\delta_n}^{\alpha=0}(Z^n_1)$. Same setting and scenarios as the ones presented in Figure \ref{fig1}, however in this case  $w\in  \left\{0.0005, 0.001, 0.005, 0.01, 0.05, 0.1, 0.5, 1 \right\}$ with $l=0.167$ and $\alpha=0$.}
\label{fig2}
\end{figure}

Similarly, Figure \ref{fig2} presents the expressiveness of the scaling factor $w$ in $b_n$ and its effects on the estimation of MI in the two aforementioned regimes $\sigma=0$ and $\sigma=0.7$.  For these plots,  we fixed $l=0.167 \in (0.133,1/3)$ and $\alpha=0$, i.e., we focus on the performance of the  full tree. In the finite sampling regime,  $n\in \left\{1,..,10^5 \right\}$, the effect of $w$ is relevant in terms of the capacity to over-estimate $I(X,Y)$ under $\sigma=0 \, (\mathcal{H}_0)$ and under-estimate or over-estimate $I(X,Y)$ under the case $\sigma=0.7$. The effect of $w$ in the performance of the full-tree solution is more prominent than the effect of $l$ in both scenarios ($\mathcal{H}_0$ and $\mathcal{H}_1$).

\begin{figure}
\centering
\includegraphics[width=1.00\textwidth]{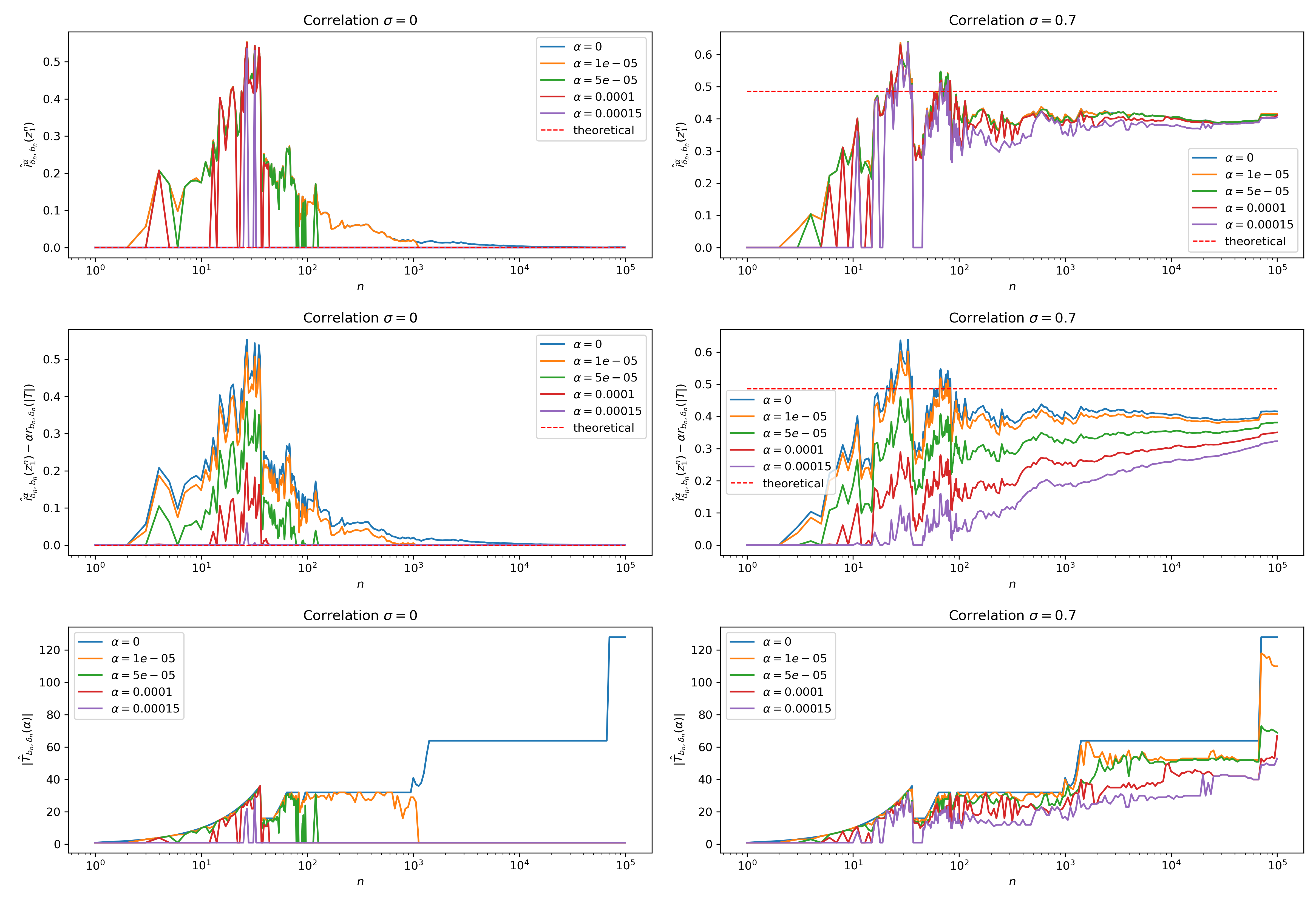}
\caption{Examples of $\hat{i}_{b_n,\delta_n}^{\alpha}(Z^n_1)$,  $\tilde{i}_{b_n,\delta_n}(\alpha)$ and $\left|  \hat{T}_{b_n,\delta_n} (\alpha) \right|$ function of the sampling size $n$ and for different values of $\alpha$ with $l=0.167$ and $w=0.05$. The case $\alpha=0$ corresponds to the non-regularize solution (the full tree).}
\label{fig3}
\end{figure}

Finally, we look at the effects of the scalar $\alpha$ in {\bf Eq.(24)}, 
for which we select $l=0.167\in (0.133,1/3)$ and $w=0.05\in (0.005, 0.1)$.  Figure \ref{fig3} shows different selections for this factor in the two scenarios already introduced ($\sigma=0$ and $\sigma=0.7$)  and  $n\in \left\{1,.., 10^5\right\}$. We present three sets of figures: one plotting the plug-in estimations $\hat{i}_{b_n,\delta_n}(\alpha) \equiv D_{\sigma({\pi_{b_n,\delta_n}^\alpha})}(\hat{P}_n||\hat{Q}_n^*) $, the other the regularized MI estimations $\tilde{i}_{b_n,\delta_n}(\alpha) \equiv  \hat{i}_{b_n,\delta_n}(\alpha) - \alpha \cdot r_{b_n, \delta_n}(\left| \hat{T}_{b_n,\delta_n} (\alpha) \right|)$ and finally, the size of the optimal tree: $ \left|  \hat{T}_{b_n,\delta_n} (\alpha) \right|$.  In general, we observe that the best results under $\mathcal{H}_1$ are obtained when $\alpha$ is very small, in the order of $\sim 10^{-5}$. This finding is interesting because it shows that, in general, the theoretical confidence interval that is used to obtain the regularization term in {\bf Eq.(16)}  is very conservative. This means that in the non-asymptotic regime, this term needs to be corrected to obtain good results. It is worth noticing that this scaling factor $\alpha$ is fixed (independent of $n$) and it does not affect the theoretical properties of our scheme: in particular, the structural capacity of our TSP scheme to detect independence under $\mathcal{H}_0$ presented in {\bf Section VI}. This capacity is observed in Figure \ref{fig3} that present the size of $ \hat{T}_{b_n,\delta_n} (\alpha)$. Here, the trivial tree (with one cell) is obtained in a finite number of steps as predicted by our results in {\bf Section VI.A}. On the other hand, for the regime of high MI, we observe that the tree's size is comparable to the size of the full-tree (obtained for $\alpha=0$). Importantly, both of these results support the adaptive nature of our data-driven solution.

\subsection{Complementing Results for {\bf Figure 3}}
\label{sec_comp}
For selecting parameters of the alternative methods (the $L_1$ test, the log-likelihood test, and the Pearson-$\chi^2$ test), we implemented them and explored their performances in the admissible range where those approaches are known to be strongly consistent \cite{gretton_2010}. In particular,  we consider the number of partitions per dimension as $m(n)=n^p$ with $p\in(0,0.5)$ (details in \cite{gretton_2010}), which satisfies the strong-consistency conditions for the $L_1$ and the log-likelihood tests established in \cite{gretton_2010}. Although there is no explicit proof of Pearson-$\chi^2$ test strong-consistency, we use a regime of values for $p \in (0,1/3)$ according to the range suggested in \cite{gretton_2010}.  

\begin{figure} 
\centering
\includegraphics[width=0.95\textwidth]{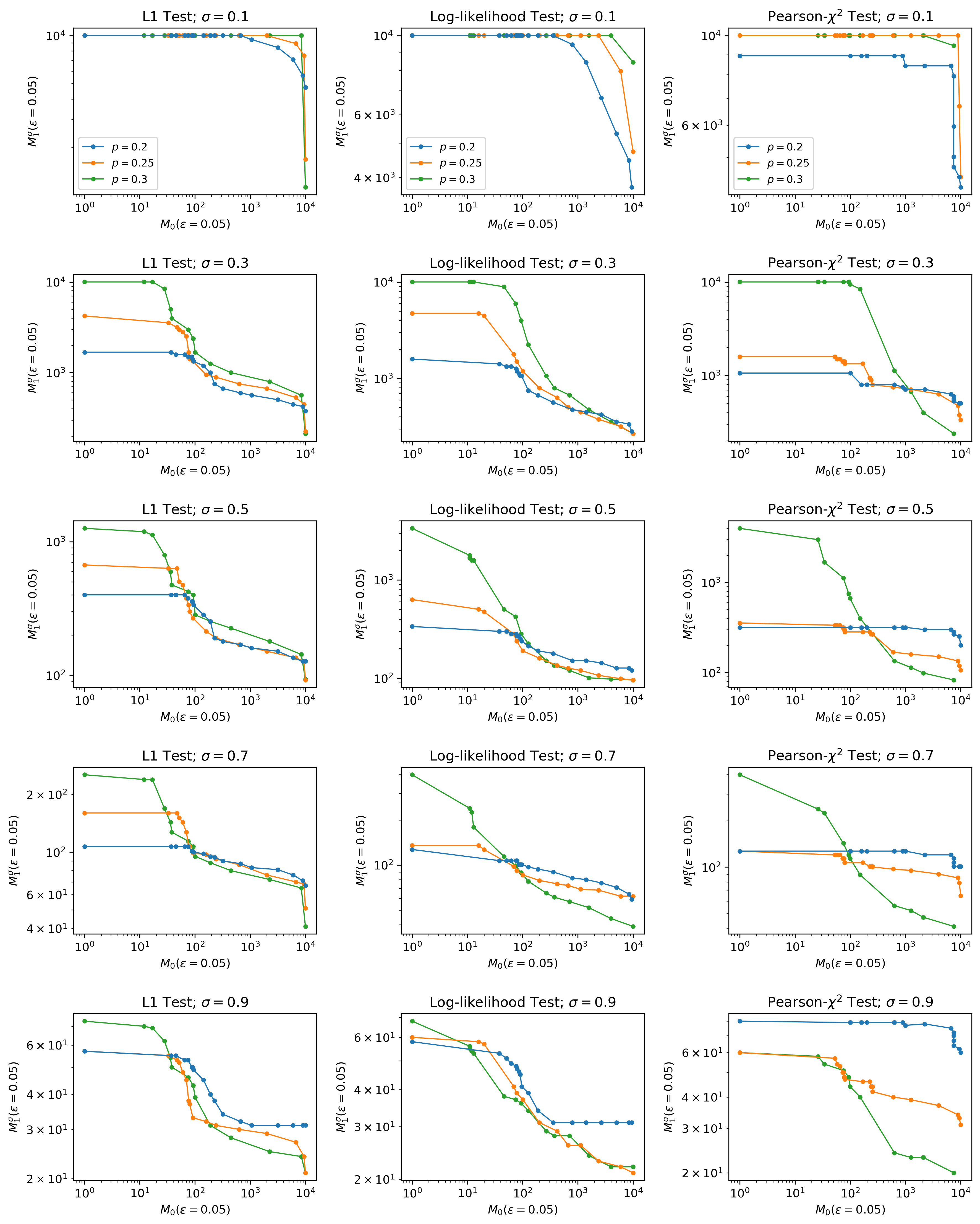}
\caption{Illustration of the trade-off between $M_0(\epsilon)$ and $M^\sigma_1(\epsilon)$ obtained for the $L_1$-test, the log-likelihood test and the Pearson-$\chi^2$ test, considering the number of partition for the domain of each random variable as $m(n)=n^p$, for $p\in\{0.2,0.25,0.3\}$. Five scenarios of correlation are presented for $\mathcal{H}_1$ considering $\sigma = \mathbb{E}(XY) \in \left\{0.1, 0.3,0.5,0.7, 0.9 \right\}$.}
\label{fig7}
\end{figure}

Under the experimental setting presented in {\bf Section VIII}, Figure \ref{fig7} reports different performance curves (the trade-off between $M_0(\epsilon)$ and $M^\sigma_1(\epsilon)$) when selecting different parameters of these methods. From these curves, the selected parameters for each method were the ones that offer a smooth transition between their sampling complexity pairs trade-off. This criterion is consistent with what we did to select the parameters of our method: please refer to {\bf Figure 1}. Interestingly, for the three methods, this was obtained consistently for $p$ in the range of $0.2$. Importantly, this selection offers reasonable solutions on all the regimes of correlation presented in Figure \ref{fig7}.

\subsection{Testing the Method on Different Distributions}
\label{sec_add_exp}
To complement the experimental results in {\bf Section VIII.A} (obtained for Gaussian distributions in the scalar case), we explore the performance of our method against non-adaptive histogram-based tests by studying the detection-time trade-off curves $(M_0(\epsilon),M_1^\sigma(\epsilon))$ on data samples generated by two important family of distributions: a heavy-tailed distribution (t-student law) and a multi-modal distribution.

\begin{figure}[ht]
\centering
\includegraphics[width=0.9\textwidth]{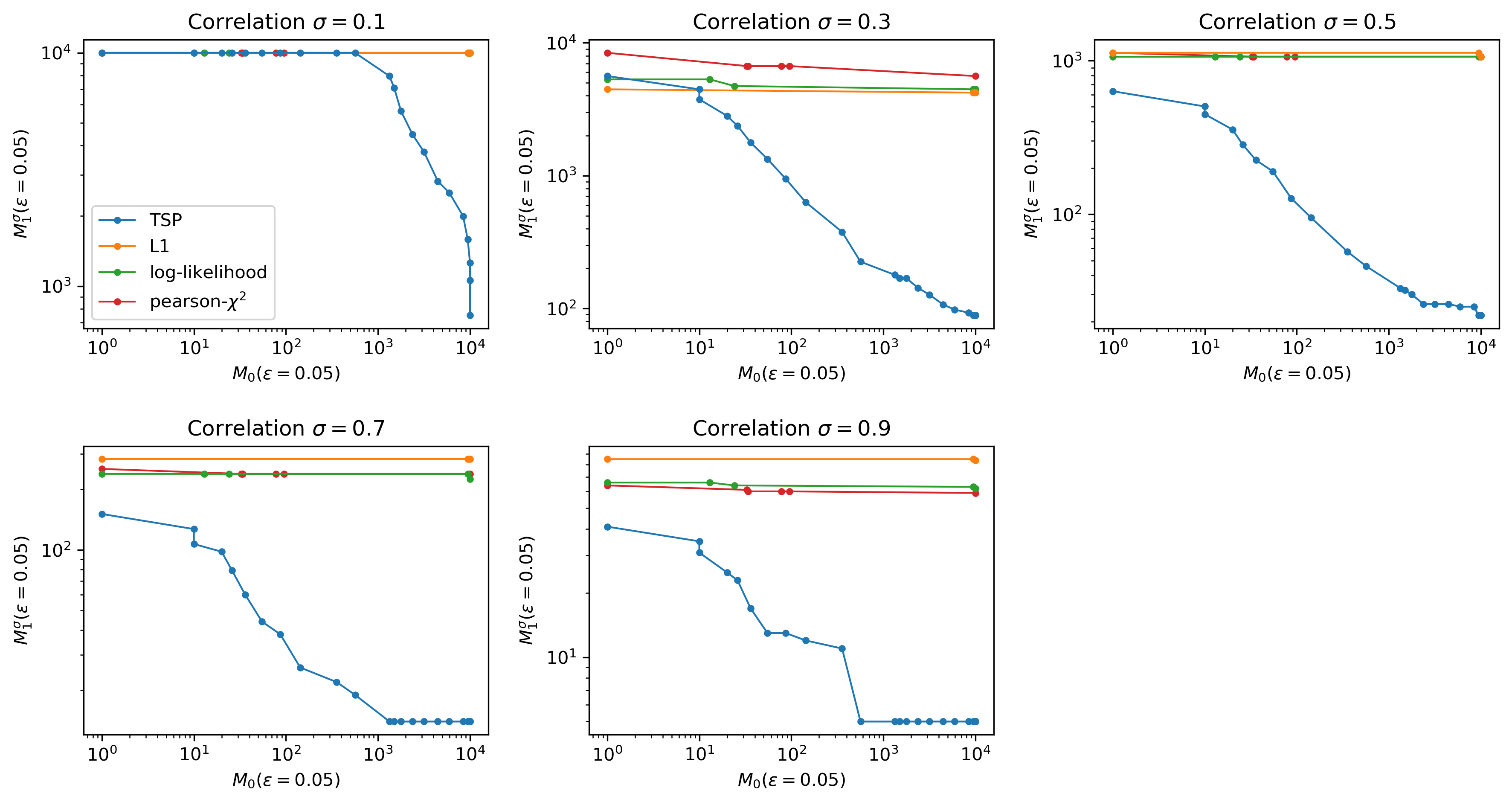}
\caption{
Illustration of the trade-off between $M_0(\epsilon)$ and $M^\sigma_1(\epsilon)$ obtained for the TSP test, the $L_1$-test, the log-likelihood test, and the $\chi^2$-test using samples from a t-student distribution with 2 degrees of freedom. The TSP test parameter configuration is $l=0.001$ and $w=0.1$, with $\alpha\in[10^{-5},5.5\cdot 10^{-4}]$. The $L_1$ test parameter configuration is $m(n)=n^{0.3}$, with $C\in[1.53,1.547]$. The log-likelihood test parameter configuration is $m(n)=n^{0.3}$, with $C\in[0.465,0.485]$. The Pearson-$\chi^2$ test parameter configuration is $m(n)=n^{0.3}$, with $C\in[0.106,0.12]$. Five scenarios of correlation are presented for $\mathcal{H}_1$ considering $\sigma = \mathbb{E}(XY) \in \left\{0.1, 0.3,0.5,0.7,0.9 \right\}$.}
\label{fig8}
\end{figure}

Figure \ref{fig8} presents the trade-off  between $(M_0(\epsilon),M_1^\sigma(\epsilon))$ for different values of correlation (under $\mathcal{H}_1$) using samples generated by a {\em t-student distribution} with $2$ degrees of freedom (which is a heavy-tailed distribution). The curves show that the performance of our test is superior to the other methods in all correlation scenarios indexed by $\sigma$. The performance gain is significant in the scenarios of high correlation and a for relatively high independence detection time (i.e., when $M_0(\epsilon)\geq 10^2$). For the non-adaptive tests, it is intriguing the relatively flat (insensitive) trade-off observed in all the explored scenarios, where the curves are close to an horizontal line. This flat trade-off behaviour in $(M_0(\epsilon),M_1^\sigma(\epsilon))$ is observed even when considering a significantly number of cells  (using $p=0.3$). In fact, this selection ($p=0.3$) produces a number of cells that is significantly higher than the number of cells obtained with our adaptive method (more of this analysis below). Therefore, this observation shows that our scheme with fewer data-driven cells produces better performance trade-off in general. In conclusion, these performance curves show the benefits of the adaptive partitioning in our test, which can deal with heavy-tailed distributions significantly better than tests that use non-adaptive partitions (binning) of the space.

\begin{figure}[ht]
\centering
\includegraphics[width=0.9\textwidth]{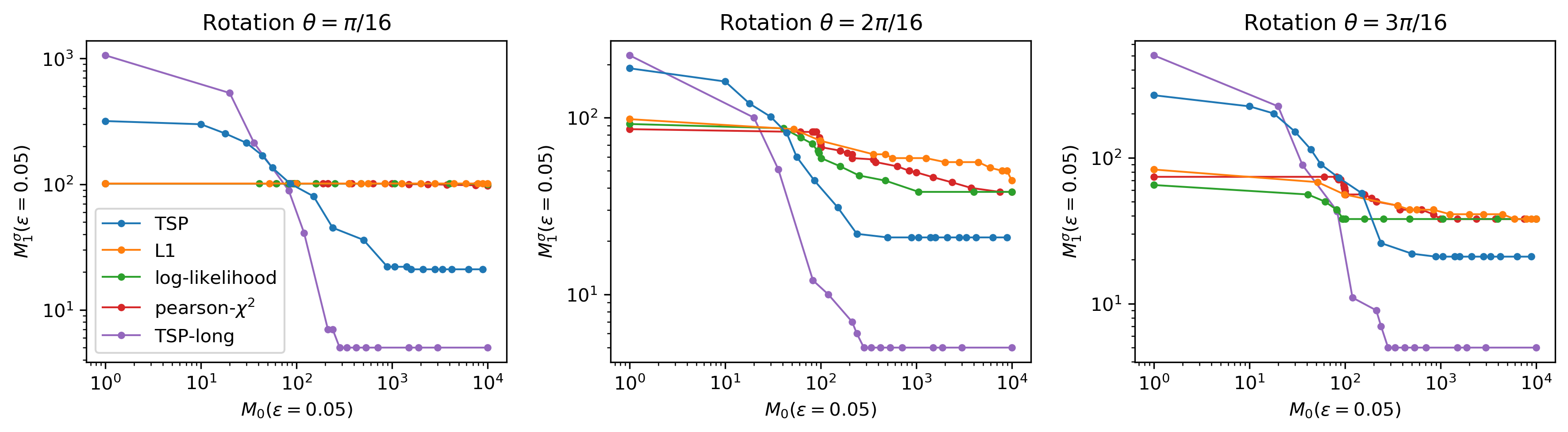}
\caption{
Illustration of the trade-off between $M_0(\epsilon)$ and $M^\sigma_1(\epsilon)$ obtained for two parameters configuration of the TSP test, the $L_1$-test, the log-likelihood test, and the $\chi^2$-test using samples from a mixture of four non-correlated Gaussians centered in $(1,1),(1,-1),(-1,1),(-1,-1)$ respectively, with $\sigma^2=0.05$. The TSP test parameter configuration is $l=0.001$ and $w=0.1$, with $\alpha\in[0.00001,0.0005]$. The TSP-long configuration is $l=0.167$ and $w=0.05$, with $\alpha\in[0.000009,0.00025]$. The $L_1$ test parameter configuration is $m(n)=n^{0.3}$, with $C\in[0.6,1.1]$. The log-likelihood test parameter configuration is $m(n)=n^{0.3}$, with $C\in[0.067,0.32]$. The Pearson-$\chi^2$ test parameter configuration is $m(n)=n^{0.3}$, with $C\in[0.075,0.0125]$. Three scenarios of rotation are presented for $\mathcal{H}_1$ considering $\theta\in \left\{\pi/16,2\pi/16,3\pi/16 \right\}$.}
\label{fig9}
\end{figure}

Figure \ref{fig9}, on the other hand, presents the trade-off between $M_0(\epsilon)$ and $M^\sigma_1(\epsilon)$ using samples generated by a mixture of four Gaussians (a multi-modal distribution). Following the setting introduced in Gretton (2008), the dependence level is produced by rotating in a counter close-wise manner (with a magnitude $\theta$) a 2D random vector with the following Gaussian mixture density: 
\begin{equation}
f(x,y) = \frac{1}{4} \sum_{i=1}^4 \mathcal{N}((x,y);\mu_i, \bar{\Sigma}),
\end{equation}
where $\mathcal{N}((x,y);\mu, \Sigma)$ denotes the bi-variate Gaussian density function with mean $\mu$ and covariance matrix $\Sigma$, evaluated in $(x,y)$. In this case, $\bar{\Sigma} = \begin{bmatrix} 0.05 & 0 \\ 0 & 0.05 \end{bmatrix}$, $\mu_1 = (1,1)$, $\mu_2 = (1,-1)$, $\mu_3 = (-1,1)$, $\mu_4 = (-1,-1)$. The independent scenario is obtained by choosing $\theta=0$, while three different dependent scenarios ($\mathcal{H}_1$) are explored by choosing $\theta\in\{\pi/16,2\pi/16,3\pi/16\}$. Similarly to the results presented in \textbf{Figure 3} (for the Gaussian case), we observed that our TSP test outperforms the other tests in the regime of $M_0(\epsilon)\geq 10^2$, while the other tests presents a better performance in the opposite regime. It is important to notice that, unlike the Gaussian case, the number of partitions of each alternative test is determined by $p=0.3$, which induces a notably higher number of cells than the ones obtained by the parameter configuration of our TSP test. It is noteworthy that a value of $p=0.3$ induces $m(n)^2=\left(\left \lfloor n^p\right \rfloor \right)^2
=\left(\left \lfloor 1000^{0.3}\right \rfloor \right)^2=49$ partitions for the alternative tests, while $l=0.001$ and $w=0.1$ induces just $8$ partitions for the tree without pruning, i.e., the final number of partitions will be lower than $8$. This demonstrates the efficiency of the proposed method, that can represent the statistical dependence of random variable with a considerably lower number of cells because of its adaptive capacity. Due to the multi-modal  nature of the distribution used in these experiments, we present an alternative configuration of a TSP test with $l=0.167$ and $w=0.05$ (denoted as TSP-long in Figure \ref{fig9}),inducing more adaptive cells (but still lower than the other tests) in order to capture the statistical dependence of the multi-modal distribution in a less conservative way. We observed a considerably benefit in this new choice in the regime when $M_0(\epsilon)\geq 10^2$ with a moderate-slight deterioration in the opposite regime.

\begin{figure}[ht]
\centering
\includegraphics[width=1.00\textwidth]{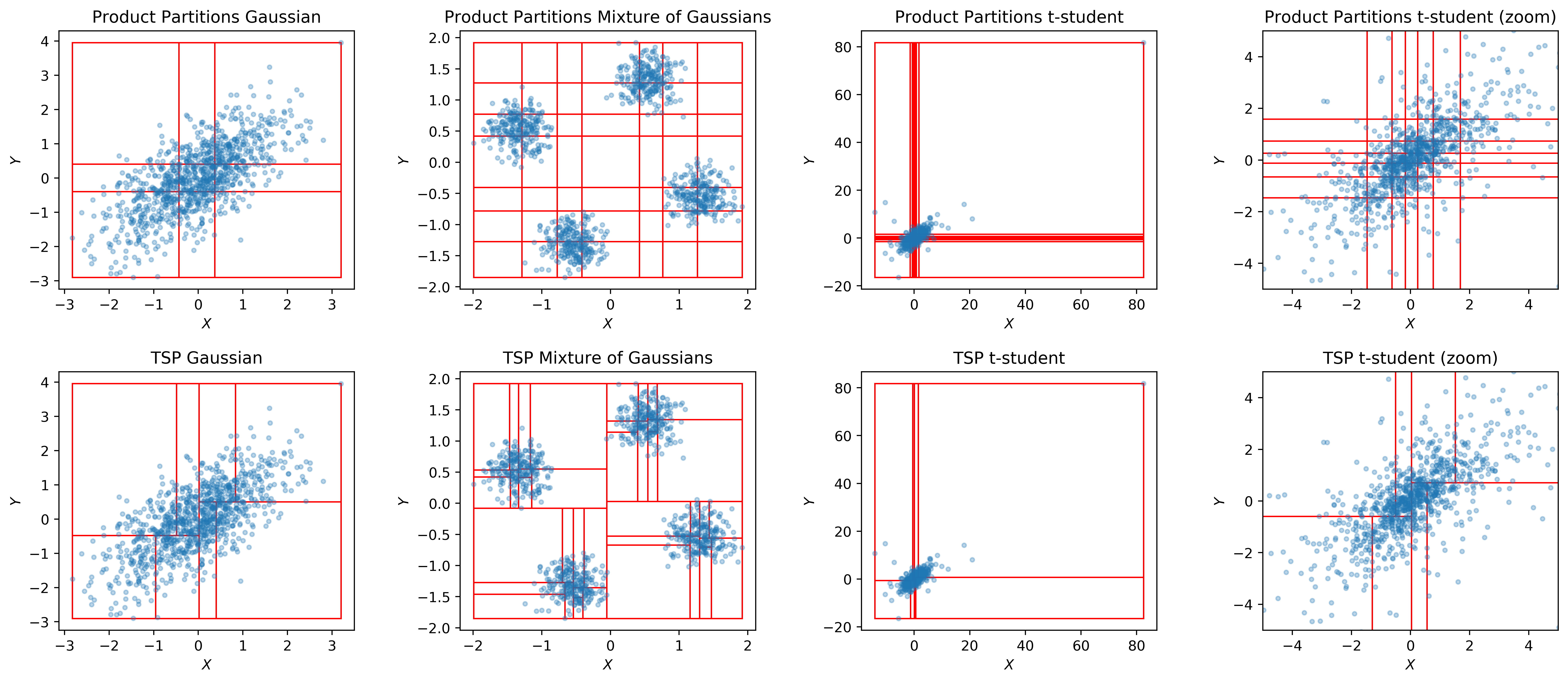}
\caption{
Illustration of partitions obtained from the the non-adaptive product partitions scheme (first row) and TSP adaptive partition scheme without pruning (second row), for $n=1000$ samples of a simply Gaussian distribution, a mixture of four Gaussians distribution and a t-student distribution. The parameters of the non-adaptive product partition are $p=0.2$ for the Gaussian distribution and $p=0.3$ for the mixture of Gaussians and t-student distributions. The parameters of the TSP partition are $l=0.001,w=0.1$ for the Gaussian and t-student distribution and $l=0.167,w=0.05$ (long tree) for the mixture of Gaussian distributions.}
\label{fig10}
\end{figure}

To complement this analysis,
Figure \ref{fig10} shows the partitions generated by the non-adaptive and our adaptive partition schemes. For this, $n=1000$ samples were considered for each of the three distributions explored (Gaussian, t-student and Gaussian Mixture). The adaptive attribute of our scheme is observed and contrasted with non-adaptive partitions, which is particularly evident in the multi-modal case.  We observe that our scheme, as expected, represent the distribution of samples 
more effectively (with a lower number of cells) than non-adaptive scheme. Indeed, non-adaptive partitions use a considerably higher number of cells to represent the statistical dependence observed in the data, which can be used to explain the difference in performance with our method. In particular, there is a clear issue observed for the non-adaptive scheme: most of the cells have almost no samples, which is an ineffective way to represent the information of the data. This could significantly affect the discrimination ability of this non-adaptive methods for small sample-size (see Figure \ref{fig10}). From our perspective,  this is one of the primary sources to explain the erratic (insensitive) flat behaviour observed in many of the trade-off curves presented in Figs \ref{fig8} and \ref{fig9}, and overall, the better (and more expresive) performance trade-off offered by of our approach.

\subsection{Statistical Significance Analysis in Section IX.C.}
\label{app_stat_sig}

\subsubsection{Proof of Proposition 2}

\begin{proof}
Let us consider the set 
\begin{equation}\label{eq_app_stat_sig_1}
	A_{b_n,\delta_n,a_n} \equiv  \left\{z^n=(x^n,y^n) \in {\mathcal X}^n \times {\mathcal Y} ^n: \phi_{b_n,\delta_n,a_n}, (z^n)=0 \right\}, 
\end{equation}
i.e., the points in the space where our test of length $n$ decides $\mathcal{H}_0$. Under $\mathcal{H}_0$, the probability of rejecting the null is given by: 
\begin{equation}\label{eq_app_stat_sig_2}
	\alpha(b_n,\delta_n,a_n) = \mathbb{P}_{Z^n}( A^c_{b_n,\delta_n,a_n}), 
\end{equation}
where $\mathbb{P}_{Z^n}$ denotes the probability of $Z^n$ induced by $P$ (i.e., the $n$-fold product distribution).
Using the argument presented in the proof of {\bf Theorem 3} (see {\bf Eq.(51)}), we show that when our data-driven partition $\pi_{b_n,\delta_n}()$ achieves the trivial case $\left\{R^d \right\}$ (the partition with one cell), then $\phi_{b_n,\delta_n,a_n}(\cdot)$ decides $0$, independent of the value of $a_n>0$. 
This {\em structural detection of independence} (see {\bf Section VI.A} for more details) implies that:
\begin{equation}\label{eq_app_stat_sig_3}
		B_{b_n,\delta_n} \equiv  \left\{z^n=(x^n,y^n): \pi_{b_n,\delta_n}(z^n)= \left\{R^d \right\} \right\} \subset A_{b_n,\delta_n,a_n},  
\end{equation} 
because $\pi_{b_n,\delta_n}(z^n) = \left\{R^d \right\}$ implies that $\phi_{b_n,\delta_n,a_n}(z^n)=0$ independent of the magnitude of  $a_n>0$. 

At this point we use the assumptions of {\bf Theorem 2}. In the argument presented in the proof of {\bf Theorem 2}, 
it was proved that (see {\bf Appendix III}) the set $\xi^{n}_{\delta_n,b_n}$ (see {\bf Eq.(39) in  Appendix III}) satisfies that if $z^n\in \xi^{n}_{\delta_n,b_n}$ then $z^n\in B_{b_n,\delta_n}$ (i.e., $\pi_{b_n,\delta_n}(z^n)= \left\{R^d \right\}$). Consequently, we have that: 
\begin{equation}\label{eq_app_stat_sig_4}
	\xi^{n}_{\delta_n,b_n} \subset B_{b_n,\delta_n}.
\end{equation} 
Furthermore, the proof of {\bf Theorem 2} in {\bf Appendix III} shows that ${\mathcal P}_{Z^n} (\xi^{n}_{\delta_n,b_n})\geq 1 -\delta_n$, where this bound is distribution-free (independent of $P\in \mathcal{P}_0$) from the use of the distribution-free concentration inequality in {\bf Lemma 1}  (see the argument of this in {\bf Appendix III}). 
Therefore, from (\ref{eq_app_stat_sig_3}) and (\ref{eq_app_stat_sig_4}), we conclude that
\begin{equation}\label{eq_app_stat_sig_5}
	\alpha(b_n,\delta_n,a_n) \leq {\mathcal P}_{Z^n}(B^c_{b_n,\delta_n}) \leq 1- {\mathcal P}_{Z^n}(\xi^{n}_{\delta_n,b_n}) \leq \delta_n, 
\end{equation}
where this upper bound is valid for any model $P\in \mathcal{P}_0$.
\end{proof}

\subsubsection{General Connection with Strong Consistency}
There is a connection between strong consistency (in {\bf Definition 1}) and the statistical significance (and also the power of a test) that can be obtained from the proof of {\bf Theorem 1} in {\bf Appendix I}. Using the notation of {\bf Theorem 1}, if we have a general test $(\phi_n(\cdot))_{n\geq 1}$, 
	the test is consistent if under ${\mathcal H}_0$, the binary process $(\phi_n(Z^n))_{n\geq 1}$ reaches $0$ eventually
	almost surely, which is equivalent to state that: 
	\begin{equation}\label{eq_app_stat_sig_6}
		\mathbb{P}(\lim\sup_n A_n^c)= \mathbb{P}(\cap_{m \geq 1} \cup_{n \geq m} A_n^c)
								 =\lim_{m \rightarrow \infty} \mathbb{P}(\cup_{n \geq m} A_n^c)=0
	\end{equation}
	where $A_n \equiv \left\{z^n\in {\mathcal X}^n \times {\mathcal Y}^n, \phi_n(z^n) =0 \right\}$. Using this notation,
	the statistical significance of $\phi_n()$ under ${\mathcal H}_0$ is $\alpha_n \equiv \mathbb{P}_{Z^n}(A_n^c)$.  
	Therefore, under ${\mathcal H}_0$, we have that: 
	\begin{equation}\label{eq_app_stat_sig_7}
		\alpha_n=\mathbb{P}_{Z^n}(A_n^c) \leq  \mathbb{P}(\cup_{k \geq n} A_k^c).
	\end{equation}
	Therefore, if the test is strongly consistent from (\ref{eq_app_stat_sig_6}) and the bound in  (\ref{eq_app_stat_sig_7}), 
	it follows that the statistical significance of the test tends to zero for any model where $X$ and $Y$ are independent ($P \in {\mathcal P}_0$).

	Concerning the power, under ${\mathcal H}_1$ we have that the binary process $(\phi_n(Z^n))_{n\geq 1}$ reaches $1$ eventually almost surely (under the assumption of consistency), which is equivalent to state that: 
	\begin{equation} \label{eq_app_stat_sig_8}
		\mathbb{P}(\lim\inf_n A_n^c)= \mathbb{P} (\cup_{m \geq 1} \cap_{n \geq m} A_n^c)
								 =\lim_{m \rightarrow \infty} \mathbb{P}(\cap_{n \geq m} A_n^c)=1. 
	\end{equation}
	 Therefore using (\ref{eq_app_stat_sig_8}), the power of the test $\beta_n = \mathbb{P}_{Z^n}(A_n^c) \geq \mathbb{P} (\cap_{k \geq n} A_k^c) \rightarrow 1$ as $n$ tends to infinity from (\ref{eq_app_stat_sig_8}), for any model where $X$ and $Y$ are not independent ($P \in {\mathcal P}_1$).
	  
	 In conclusion, the definition of consistency used in this work ({\bf Definition 1}) implies a vanishing error 
	 (when $n$ tends to infinity) under both hypotheses. However,  this asymptotic property does not offer specifics bounds for the errors when $n$ is finite, in contrast to the result shown in Eq.(\ref{eq_app_stat_sig_5}).

\end{document}